\documentclass[11pt,journal,compsoc,onecolumn,]{IEEEtran}
\usepackage{natbib} 
\usepackage{url} % This package helps with URLs in citations % Set the citation style to APA 
\makeatletter
\def\thickhline{%
	\noalign{\ifnum0=`}\fi\hrule \@height \thickarrayrulewidth \futurelet
	\reserved@a\@xthickhline}
\def\@xthickhline{\ifx\reserved@a\thickhline
	\vskip\doublerulesep
	\vskip-\thickarrayrulewidth
	\fi
	\ifnum0=`{\fi}}
\makeatother

\newlength{\thickarrayrulewidth}
\usepackage{epigraph}
\usepackage{longtable}
\usepackage{setspace}

\usepackage{textcomp}

\usepackage[hidelinks]{hyperref}
\hypersetup{
	colorlinks=true,
	linkcolor=black,
	filecolor=black,
	citecolor=black,
	urlcolor=blue,
}

\usepackage{multirow}

\usepackage{color}
\usepackage[table, dvipsnames]{xcolor}
\definecolor{PyOrange}{RGB}{255, 201, 14}
\definecolor{PyBlue}{RGB}{112, 146, 190}

\definecolor{WordGreen}{RGB}{100, 136, 40}
\definecolor{WordDarkGrey}{RGB}{82, 82, 82}
\definecolor{WordRed}{RGB}{192, 80, 77}
\definecolor{WordBlue}{RGB}{0, 122, 192}

\definecolor{WordLightBlue}{RGB}{218, 238, 243}
\definecolor{WordLightGreen}{RGB}{234, 241, 221}

\definecolor{WordFillGreen}{RGB}{194, 214, 155}
\definecolor{WordFillRed}{RGB}{252, 214, 182}
\definecolor{WordFillGray}{RGB}{217, 217, 217}

\UseRawInputEncoding
\usepackage{times}
\usepackage{epsfig}
\usepackage{amssymb}
\usepackage[ruled,vlined,linesnumbered]{algorithm2e}
\usepackage{comment}
\usepackage{array, makecell} 
\usepackage{lettrine}
\usepackage{url} 
\usepackage{lscape}
\usepackage{bbm,bm}
\usepackage{tabularx}
\usepackage{colortbl}
\usepackage{hhline}
\usepackage{vcell}
\usepackage{longtable}
\usepackage{vcell}
\usepackage{rotating}
\usepackage{pdflscape}
\usepackage{sidecap}
\usepackage{neuralnetwork}
\usepackage[export]{adjustbox} % https://tex.stackexchange.com/questions/91566/syntax-similar-to-centering-for-right-and-left
\usepackage{acronym}
\acrodef{FCN}[FCN]{Fully Convolutional Network}
\acrodef{GAME}[GAME]{Grid Average Mean Absolute Error}
\acrodef{DL}[DL]{Deep Learning}
\acrodef{DNN}[DNN]{Deep Neural Network}
\acrodef{ML}[ML]{Machine Learning}
\acrodef{CV}[CV]{Computer Vision}
\acrodef{AI}[AI]{Artificial Intelligence}
\acrodef{CNN}[CNN]{Convolutional Neural Network}
\acrodef{RNN}[RNN]{Recurrent Neural Network}
\acrodef{GAN}[GAN]{Generative Adversarial Network}
\acrodef{JCU}[JCU]{James Cook University}
\acrodef{MAE}[MAE]{Mean Average Error}
\acrodef{MAP}[mAP]{Mean Average Precision}
\acrodef{CA}[CA]{Classification Accuracy}
\acrodef{LCFCN}[LCFCN]{Localization-based Counting loss Fully Convolutional Network}
\acrodef{IoT}[IoT]{Internet of Things}
\acrodef{MLP}[MLP]{Multi-Layer Perceptrons}
\acrodef{RAFT}[RAFT]{Recurrent All-Pairs Field Transforms }
\usepackage{xspace}

\usepackage[capitalize]{cleveref}
\crefname{section}{Sec.}{Section}
\usepackage{booktabs}
\usepackage{arydshln}
\usepackage{listings}
\definecolor{codegreen}{rgb}{0,0.6,0}
\definecolor{codegray}{rgb}{0.5,0.5,0.5}
\definecolor{codepurple}{rgb}{0.58,0,0.82}
\definecolor{backcolour}{rgb}{0.95,0.95,0.92}
\definecolor{cleacolorr}{rgb}{1,1,1}

\lstdefinestyle{mystyle}{
    backgroundcolor=\color{cleacolorr},   
    commentstyle=\color{codegreen},
    keywordstyle=\color{magenta},
    numberstyle=\tiny\color{codegray},
    stringstyle=\color{codepurple},
    basicstyle=\ttfamily\footnotesize,
    breakatwhitespace=false,         
    breaklines=true,                 
    captionpos=b,                    
    keepspaces=true,                 
    numbers=left,                    
    numbersep=5pt,                  
    showspaces=false,                
    showstringspaces=false,
    showtabs=false,                  
    tabsize=2
}

\usepackage[most]{tcolorbox}
\newtcolorbox[auto counter]{pabox}[2][]{%
colback=blue!5!white,colframe=blue!75!black,fonttitle=\bfseries,
title=Box~\thetcbcounter: #2,#1}
\newcommand{\MyPaperTitle}{Adaptive Deep Learning Framework for Robust Unsupervised Underwater Image Enhancement}

\usepackage{tikz}

\ifCLASSOPTIONcompsoc
  \usepackage[caption=false,font=normalsize,labelfont=sf,textfont=sf]{subfig}
\else
  \usepackage[caption=false,font=footnotesize]{subfig}
\fi

\usepackage{graphicx}
\graphicspath{{./Graphics/}}
\DeclareGraphicsExtensions{.pdf,.jpeg,.png,.jpg}

\usepackage{amsmath}

\usepackage{array}

\usepackage{url}
\usepackage[switch]{lineno}

\begin{document}

% \linenumbers  %xxxxxxxxxxxxxxxxxxxxxxxxxxxxxxxxxxxx

% \begin{titlepage}
% \begin{center}
% \vspace*{1cm}

% \textbf{ \large Adaptive Uncertainty Distribution in Deep Learning for Unsupervised Underwater Image Enhancement}

% \vspace{1.5cm}

% % Author names and affiliations
% Alzayat Saleh$^a$ (alzayat.saleh@my.jcu.edu.au), 
% Marcus Sheaves$^a$ (marcus.sheaves@jcu.edu.au), 
% Dean Jerry$^{a,b}$ (dean.jerry@jcu.edu.au), 
% Mostafa~Rahimi~Azghadi$^{a,b}$ (mostafa.rahimiazghadi@jcu.edu.au) \\

% \hspace{10pt}

% \begin{flushleft}
% \small  
% $^a$ College of Science and Engineering, James Cook University, 1 James Cook Drive, Townsville, 4811, QLD, Australia\\
% $^b$ ARC Research Hub for Supercharging Tropical Aquaculture through Genetic Solutions, James Cook University, 1 James Cook Drive, Townsville, 4811, QLD, Australia \\
% % $^c$ James Cook University, 1 James Cook Drive, Townsville, 4811, QLD, Australia

% \begin{comment}
% Clearly indicate who will handle correspondence at all stages of refereeing and publication, also post-publication. Ensure that phone numbers (with country and area code) are provided in addition to the e-mail address and the complete postal address. Contact details must be kept up to date by the corresponding author.
% \end{comment}

% \vspace{1cm}
% \textbf{Corresponding Author:} \\
% Mostafa~Rahimi~Azghadi \\
% College of Science and Engineering, James Cook University, 1 James Cook Drive, Townsville, 4811, QLD, Australia\\
% Email: mostafa.rahimiazghadi@jcu.edu.au

% \end{flushleft}        
% \end{center}
% \end{titlepage}

\title{\MyPaperTitle}

\author{
    \IEEEauthorblockN{
        Alzayat Saleh\IEEEauthorrefmark{1}\IEEEauthorrefmark{3}, 
        Marcus Sheaves\IEEEauthorrefmark{1}\IEEEauthorrefmark{3}, 
        Dean Jerry\IEEEauthorrefmark{1}\IEEEauthorrefmark{2}\IEEEauthorrefmark{3}, 
        and Mostafa~Rahimi~Azghadi\IEEEauthorrefmark{1}\IEEEauthorrefmark{2}\IEEEauthorrefmark{3}
    }
    
    \IEEEauthorblockA{\IEEEauthorrefmark{1}College of Science and Engineering, James Cook University, Townsville, QLD, Australia}
    \IEEEauthorblockA{\IEEEauthorrefmark{2}ARC Research Hub for Supercharging Tropical Aquaculture through Genetic Solutions, James Cook University, Townsville, QLD, Australia}
    \IEEEauthorblockA{\IEEEauthorrefmark{3}\{alzayat.saleh, marcus.sheaves, dean.jerry, mostafa.rahimiazghadi\}@jcu.edu.au}
}

% \affil[1]{College of Science and Engineering, James Cook University, Townsville, QLD, Australia}
% \affil[2]{ARC Research Hub for Supercharging Tropical Aquaculture through Genetic Solutions, James Cook University, Townsville, QLD, Australia}

\maketitle

\begin{abstract}
One of the main challenges in deep learning-based underwater image enhancement is the limited availability of high-quality training data. Underwater images are often difficult to capture and typically suffer from distortion, color loss, and reduced contrast, complicating the training of supervised deep learning models on large and diverse datasets. This limitation can adversely affect the performance of the model.
In this paper, we propose an alternative approach to supervised underwater image enhancement. Specifically, we introduce a novel framework called Uncertainty Distribution Network (\texttt{UDnet}), which adapts to uncertainty distribution during its unsupervised reference map (label) generation to produce enhanced output images. UDnet enhances underwater images by adjusting contrast, saturation, and gamma correction. It incorporates a statistically guided multicolour space stretch module (SGMCSS) to generate a reference map, which is utilised by a U-Net-like conditional variational autoencoder module (cVAE) for feature extraction. These features are then processed by a Probabilistic Adaptive Instance Normalisation (PAdaIN) block that encodes the feature uncertainties for the final image enhancement. The SGMCSS module ensures visual consistency with the input image and eliminates the need for manual human annotation. Consequently, UDnet can learn effectively with limited data and achieve state-of-the-art results.
We evaluated UDnet on eight publicly available datasets, and the results demonstrate that it achieves competitive performance compared to other state-of-the-art methods in both quantitative and qualitative metrics. Our code is publicly available at \href{https://github.com/alzayats/UDnet}{https://github.com/alzayats/UDnet}.
\end{abstract}

\ifCLASSOPTIONpeerreview
\else
\fi

%%%%%%%%%%%%%%%%%%%%%%%%%%%%%%%%%%%%%%%%%%%%%%%%%%%%%%%%%%%%%%%%

%%%%%%%%%%%%%%%%%%%%%%%%%%%%%%%%%%%%%%%%%%%%%%%%%%%%%%%%%%%%%%%%

%%%%%%%%%%%%%%%%%%%%%%%%%%%%%%%%%%%%%%%%%%%%%%%%%%%%%%%%%%%%%%%%

\section{Introduction}\label{secintro}

\begin{figure*}[ht]%[!t]
\centering
\includegraphics[width=0.98\textwidth]{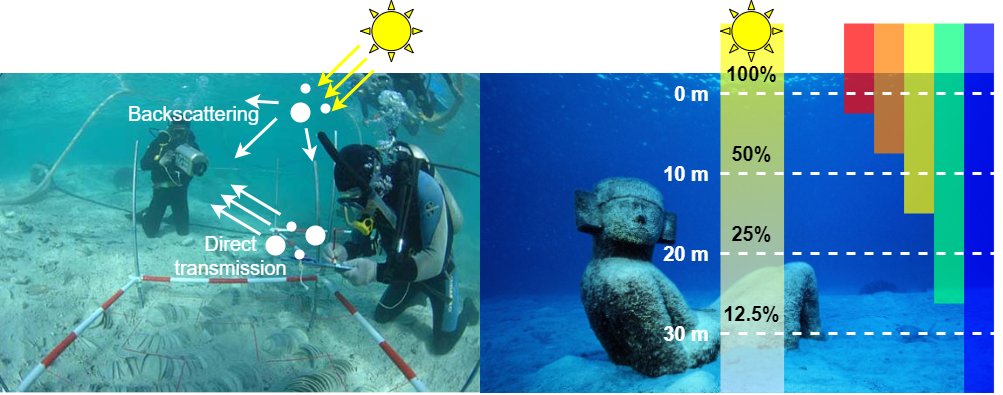}
\caption{(Left) Natural light entering the water is scattered multiple times, forming the backscattering for the underwater scene. The light directly reflected off objects in the scene also travels to the camera, and the total light perceived is the sum of these two components, creating the colours and details in underwater images. 
(Right) Different wavelengths of light are absorbed and scattered differently as they travel through water. Blue light travels the longest distance due to its shorter wavelength, making underwater objects appear blue in colour.
% \citep{Chiang2011UnderwaterDehazing}
}
\label{fig:9}
\end{figure*}

The enhancement of underwater images is a critical task in computer vision, with applications ranging from underwater robotics to marine biology. However, this task presents unique challenges due to the complex optical properties of water, such as random distortion, low contrast, and wavelength-dependent absorption \citep{ji2024dual}. These factors result in colour casts, blurriness, and uneven illumination, making underwater images inherently difficult to process and analyze, see \cref{fig:9}. Addressing these challenges is crucial for improving the accuracy and reliability of tasks like object detection and target recognition in underwater environments.

%%%%%%%%%%%%%%%%%%%%%%%%%%%%%%%%%%%%%%%%%%%%%%%%%%%%%%%%%%%%%%%
% % %%%%%%%%%%%%%%%%%%%%%%%%%%%%%%%%%%%%%%%%%%%%%%%%%%%%%%%%%%%%%%%%

Over the years, various approaches have been proposed to enhance underwater images. Traditional methods, such as histogram equalization and contrast stretching, attempt to improve image visibility by redistributing pixel intensities or enhancing specific features. While these methods are computationally efficient, they often fail to address the unique complexities of underwater environments, such as non-uniform lighting and scattering effects. In contrast, deep learning-based techniques have shown great promise, leveraging large datasets to learn complex representations for image enhancement. Supervised approaches, such as those employing U-Net architectures and generative adversarial networks (GANs) \citep{zheng2024learnable}, have achieved significant improvements in underwater image quality. However, these methods rely heavily on paired training data—underwater images and their corresponding ground truth—which are challenging and costly to acquire in underwater scenarios.

Despite these advancements, the existing methods face critical limitations. Traditional approaches lack adaptability to diverse underwater conditions, while supervised learning techniques are constrained by their dependence on annotated datasets and their potential for overfitting to specific domains. Furthermore, many deep learning methods struggle to generalize effectively to new datasets, limiting their applicability in real-world underwater environments \citep{cheng2024fdce}.

To address these limitations, we propose a novel unsupervised framework for underwater image enhancement, termed the Uncertainty Distribution Network (\texttt{UDNet}). This paper presents \texttt{UDNet}, a novel unsupervised deep learning framework for robust underwater image enhancement. Unlike supervised methods that rely on large datasets of paired raw and enhanced underwater images, \texttt{UDNet} learns without using any ground truth or manually annotated images. Instead, it leverages a probabilistic approach that embraces the uncertainty inherent in underwater image enhancement. Unlike traditional methods, \texttt{UDNet} eliminates the need for paired training data, making it more practical and scalable for underwater applications. Our method introduces adaptive enhancement through uncertainty modeling, leveraging a Statistically Guided Multi-Color Space Stretch (SGMCSS) module to generate diverse reference maps. During training, the model randomly selects from these maps, enabling it to learn robust and generalized representations of underwater environments. Additionally, \texttt{UDNet} incorporates a Probabilistic Adaptive Instance Normalization (PAdaIN) block, which enhances its ability to adapt to varying image characteristics, further improving its robustness.

While previous work, such as \citep{fu2022uiu}, has successfully demonstrated the use of uncertainty-inspired methods for underwater image enhancement, these approaches often focus on specific types of uncertainty, such as pixel-level deviations. In contrast, our proposed UDNet introduces a broader framework that models distributional uncertainty through adaptive reference selection and probabilistic feature normalization. These innovations allow our method to generalize effectively across varied underwater environments without requiring paired training data. By addressing key limitations of \citep{fu2022uiu} approach, such as its reliance on predefined uncertainty maps, our method achieves state-of-the-art performance while maintaining a lightweight and practical design for real-world applications.

The advantages of \texttt{UDNet} are evident in its performance. It demonstrates strong generalizability across multiple underwater datasets, even those it was not trained on, highlighting its applicability to a wide range of scenarios. Moreover, both qualitative and quantitative evaluations show that \texttt{UDNet} achieves state-of-the-art results, outperforming or matching existing supervised and unsupervised methods.

In summary, the contributions of this work are as follows:
\begin{itemize}
    \item We introduce \texttt{UDNet}, an unsupervised framework for underwater image enhancement that eliminates the need for paired training data, addressing a key limitation of existing methods.
    \item We propose an SGMCSS module for generating diverse reference maps, combined with a PAdaIN block for adaptive enhancement, enabling robust and generalized performance across diverse underwater conditions.
    \item We demonstrate that \texttt{UDNet} achieves state-of-the-art performance on multiple datasets, showcasing its robustness and applicability in various underwater scenarios.
\end{itemize}

The remainder of this paper is organized as follows: In \cref{secrlt}, we review related work on underwater image enhancement, highlighting the strengths and limitations of existing methods. \cref{secmethod} details the architecture of our proposed framework. In \cref{secresult}, we describe our experimental setup and present the results. Finally, we discuss our findings in \cref{secdisc} and conclude in \cref{secconc}.

%%%%%%%%%%%%%%%%%%%%%%%%%%%%%%%%%%%%%%%%%%%%%%%%%%%%%%%%%%%%%%%%

\section{Related Work} \label{secrlt}

Underwater image enhancement is a challenging and active area of research, with various approaches proposed to address issues such as low contrast, distortion, and uneven illumination \citep{liu2024zero}. These approaches can be broadly categorized into four main groups: prior-based methods, model-free methods, deep learning-based methods, and probabilistic-based methods.

\subsection{Prior-Based Methods}
Prior-based methods rely on physical models of underwater image formation to estimate the optical parameters affecting underwater images. These parameters are then reversed to reconstruct enhanced images. Examples of visual cues used in such methods include the red channel prior \citep{Huang2018Shallow-waterAcquisition}, the underwater dark channel prior \citep{Drews2013TransmissionImages}, and the underwater light attenuation prior \citep{Song2018ARestoration}. While these methods leverage physical insights, their effectiveness can be limited in highly complex underwater environments where the assumptions of the underlying models may not hold.

\subsection{Model-Free Methods}

Model-free methods enhance images without explicitly modelling the degradation process, instead focussing on improving image visibility through redistribution of pixel intensity or feature enhancement. Common techniques in this category include contrast-limited adaptive histogram equalization (CLAHE), white balance (WB), and Retinex-based methods. These approaches are computationally efficient and can be extended using fusion-based or multi-scale strategies for improved performance \citep{Drews2013TransmissionImages}.

Recently, various enhancements within this category have been proposed, including a hybrid whale optimisation algorithm designed for the enhancement of contrast \citep{braik2024hybrid} and detail in colour images and a fusion-based approach combining adaptive colour correction with improved contrast enhancement strategies for the improvement of underwater image quality \citep{raveendran2024underwater}. Other methods include a histogram equalisation model specifically developed for colour image contrast enhancement \citep{wang2024histogram}, a technique utilising interval-valued intuitionistic fuzzy sets to refine color image quality \citep{jebadass2024color}, and an intelligent underwater image enhancement method that integrates color correction with contrast stretching \citep{lei2024novel}.

These methods primarily \citep{lei2024novel} aim to improve contrast and colour balance, which are critical for underwater image enhancement. Despite their effectiveness in addressing these aspects, model-free methods often struggle to adapt to the diverse and dynamic underwater conditions. Challenges such as non-uniform lighting, scattering effects, and the varying turbidity of underwater environments remain significant limitations of these approaches. Consequently, while model-free methods are valuable for certain applications, they may benefit from integration with more adaptive or data-driven approaches to handle complex underwater imaging scenarios effectively.

\subsection{Deep Learning-Based Methods}
Deep learning-based methods utilize training data to automatically learn representations for underwater image enhancement. These methods can be divided into convolutional neural networks (CNNs) and generative adversarial networks (GANs) \citep{zheng2024learnable}. For example, CNN-based models have been developed using encoder-decoder frameworks to remove noise from underwater images, while lightweight CNN architectures incorporate scene-specific information to synthesize degraded images \citep{Wavelength}. GANs, on the other hand, have been used to generate synthetic underwater images in an unsupervised manner and to train enhancement networks using synthetic data \citep{Wang2023RCA-CycleGAN:CycleGAN}. While these methods demonstrate significant improvements, they typically rely on large paired datasets (underwater images and corresponding ground truth), which are challenging to obtain, and their performance may be limited to specific training domains.

\subsection{Probabilistic-Based Methods}
Probabilistic-based methods integrate uncertainty modeling into deep learning frameworks, providing a principled way to address disturbances, modeling errors, and uncertainties inherent in underwater environments. Conditional variational autoencoders (cVAEs) represent a notable example in this category. Variational autoencoders (VAEs) are generative models comprising an encoder that maps input data to a low-dimensional latent space and a decoder that reconstructs the data from this latent representation \citep{Kingma2019vae}. VAEs differ from traditional encoders by describing probability distributions for latent variables rather than single-point estimates, enabling them to capture diverse data characteristics. To effectively train VAEs, regularization and reconstruction losses are applied to ensure compact and meaningful representations of the input data \citep{Sohn2015LearningModels}.

VAEs and cVAEs have found applications in underwater image enhancement and related tasks. For instance, they have been used for background modeling in salient object detection \citep{Li2019Supervae:Detection}, motion sequence generation \citep{Yan2018Mt-vae:Dynamics}, and image denoising \citep{Balakrishnan2019VisualDimensions}. Recent advancements have further combined VAEs with contrastive learning to identify and enhance salient features, showcasing their versatility in various scenarios.

\citep{fu2022uiu} introduced an uncertainty-inspired underwater image enhancement framework that leverages learned uncertainty maps for image refinement. While their work effectively improves image quality by modeling pixel-level uncertainty, our approach extends this concept by incorporating adaptive uncertainty modeling through random reference selection during training. Furthermore, our Probabilistic Adaptive Instance Normalization (PAdaIN) layer aligns latent feature distributions dynamically, which enables more robust generalization across diverse underwater datasets. Unlike \citep{fu2022uiu}, who primarily focused on pixel-wise uncertainty, our method captures and leverages distributional uncertainty to achieve enhanced generalization and adaptability.

In our approach, we extend these probabilistic methods by integrating conditional VAEs with uncertainty modeling techniques, enabling robust underwater image enhancement. Our framework leverages probabilistic adaptive instance normalization to learn diverse and generalized representations of underwater environments. The experimental results demonstrate that this approach significantly improves performance across diverse datasets, addressing key limitations of existing methods.

%%%%%%%%%%%%%%%%%%%%%%%%%%%%%%%%%%%%%%%%%%%%%%%%%%%%%%%%%%%%%%%%
% % %%%%%%%%%%%%%%%%%%%%%%%%%%%%%%%%%%%%%%%%%%%%%%%%%%%%%%%%%%%%%%%%
\begin{figure*}[ht]%[!t]
\centering
\includegraphics[width=0.99\textwidth]{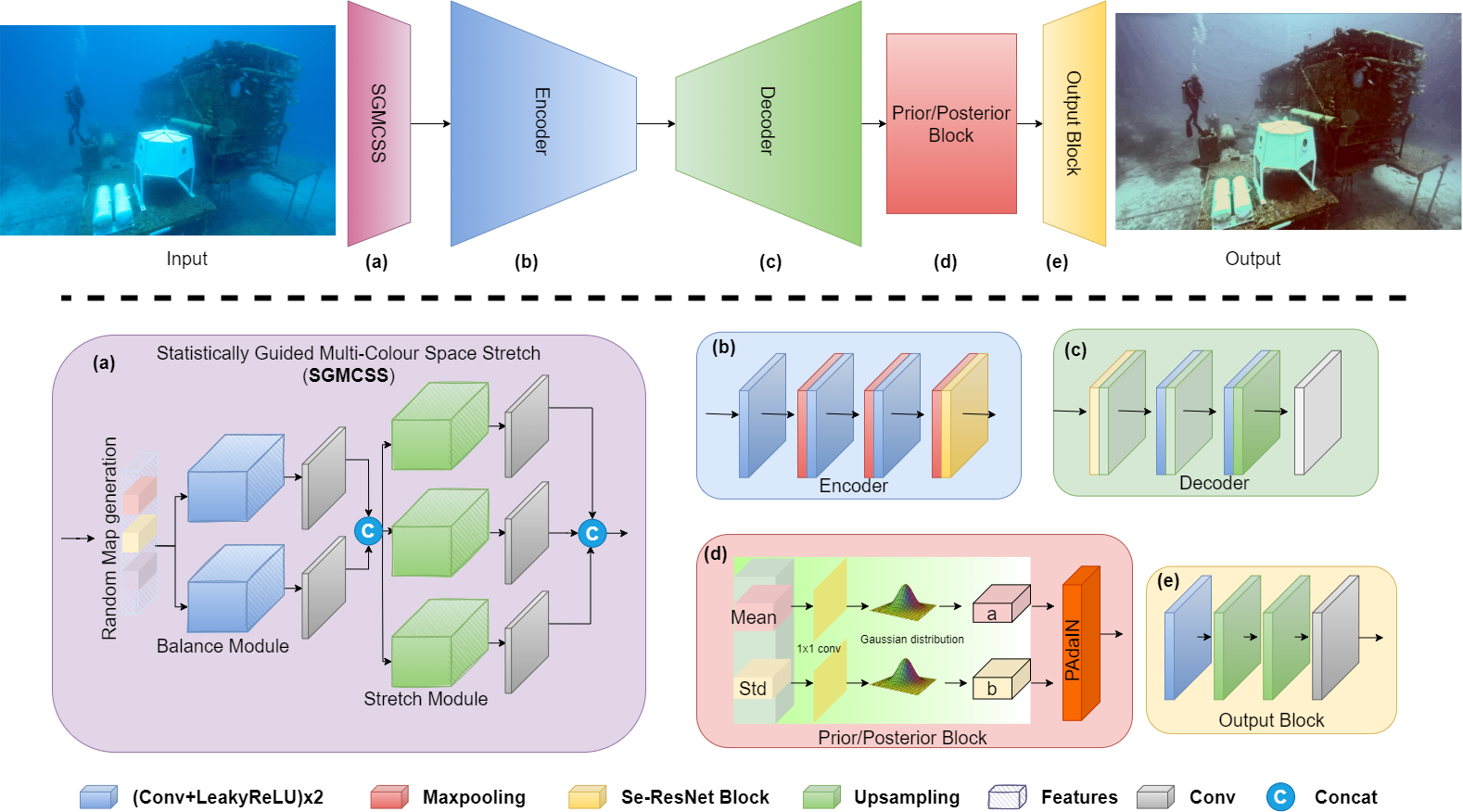}
\caption{The architecture of UDnet is composed of five primary components: (a) Statistically Guided Multi-Colour Space Stretch (SGMCSS) for reference maps generation, (b) Encoder for feature extraction, (c) Decoder for image reconstruction, (d) Prior/Posterior Block to calculate and sample Gaussian distributions for feature adaptation, and (e) Output Block that generates the enhanced underwater image. The SGMCSS module transforms the input degraded image into balanced and stretched reference maps, enabling robust feature extraction and enhancement across varying underwater image conditions. The Encoder and Decoder process the input through multi-layer convolutional operations, while the PAdaIN-enabled Prior/Posterior Block ensures stochasticity and flexibility in image enhancement.}
\label{fig:8}
\end{figure*}
% % %%%%%%%%%%%%%%%%%%%%%%%%%%%%%%%%%%%%%%%%%%%%%%%%%%%%%%%%%%%%%%%%

%%%%%%%%%%%%%%%%%%%%%%%%%%%%%%%%%%%%%%%%%%%%%%%%%%%%%%%%%%%%%%%%
\section{Method} \label{secmethod}
This section describes the various components and concepts utilized to build our Uncertainty Distribution Network (UDnet). As shown in \cref{fig:8}, UDnet is composed of three abstract building blocks including a reference map generation block that uses a statistically guided multi-colour space stretch module, a feature extractor block that uses a cVAE, and a probabilistic adaptive instance normalization block. All of these blocks and their underlying components and concepts will be discussed in detail below, however, the reader is encouraged to investigate the full detail of our implementation code at \href{https://github.com/alzayats/UDnet}{https://github.com/alzayats/UDnet}. %with the guidance of statistics to learn meaningful distributions of underwater image enhancement.
%We present the overview architecture of UDnet in \cref{fig:8}. 

\subsection{Uncertainty Distribution}

UDNet learns to adapt to the uncertainty distribution inherent in underwater image enhancement, where different images require varying degrees of enhancement in terms of contrast, saturation, gamma correction, and other factors. This uncertainty distribution refers to the inherent ambiguity that can exist in the image enhancement process, as different images need different types of enhancements. The main idea behind \texttt{UDNet} is to better incorporate this uncertainty in the enhancement process. This is motivated by the fact that the true clean image is often unavailable and that there is a degree of uncertainty in the labels used to train image enhancement models. Existing deterministic learning-guided methods \citep{Wavelength} are unable to capture this uncertainty and therefore must make compromises between different possible enhancement results.

To address this, \texttt{UDNet} employs a probabilistic framework that introduces uncertainty during the training process. Instead of relying on a fixed set of labels generated from other enhancement techniques, \texttt{UDNet} randomly selects one of three enhanced reference maps generated by applying contrast adjustment, saturation adjustment, or gamma correction to the input image. This random selection forces the model to learn a more robust and generalised representation of underwater image enhancement by accounting for the inherent ambiguity in defining the ideal enhancement.

%In order to account for the inherent uncertainty in image enhancement, UDnet introduces Uncertainty Distribution as an implicit variable that represents the various factors that could affect the outcome of the enhancement process. 
UDnet uses an implicit variable ${\bf{z}}$ to represent the uncertainty in the enhancement process. This variable could represent human subjective preferences, or the parameters of the camera or enhancement algorithms used to capture or generate the ground truth images, which could affect the outcome of the enhancement process. By taking this uncertainty into account, UDnet is able to more accurately capture the range of possible enhancements, rather than trying to determine a single "correct" result. This is particularly useful in situations where the true, unaltered image is not available or cannot be accurately reproduced.

The goal of UDNet is to learn a mapping from the low-quality input image ${\bf{x}}$ to the clean image ${\bf{y}}$ that takes into account the uncertainty represented by ${\bf{z}}$. This can be formalized as follows:
\begin{equation}\label{equ8}
p\left( {y\left| {\bf{x}} \right.} \right) \approx p\left( {{\bf{y}}\left| {{{\bf{z}}_{\max }},{\bf{x}}} \right.} \right),{{\bf{z}}_{\max }} \sim p\left( {{\bf{z}}\left| {\bf{x}} \right.} \right),
\end{equation}
where $p\left( {\bf{z}\left| \bf{x} \right.} \right)$ denotes the distribution of uncertainty, and  ${\bf{z}}_{\max }$ denotes the sample with the maximum probability.

% The equation above 
\cref{equ8} represents the probabilistic framework underlying UDnet. In this equation, $p\left( {{\bf{y}}\left| {{{\bf{z}}_{\max }},{\bf{x}}} \right.} \right)$ is the probability of the clean image ${\bf{y}}$ given the sample with the maximum probability ${\bf{z}}_{\max }$ and the low-quality input observation ${\bf{x}}$. $p({\bf{z}}|{\bf{x}})$ is the probability of the uncertainty variable given the observation. The goal of the model is to learn these probability distributions from the training data and then use them to generate enhanced images that incorporate uncertainty into the enhancement process. By doing so, UDnet is able to (1) provide users with multiple alternative enhancement results to choose from, or (2) improve the accuracy and reliability of the final enhancement result by taking the enhancement sample with the maximum probability as the final estimation, without user intervention.

 % was blue
In the proposed UDnet framework, randomness is introduced in the creation of pseudo-enhanced images through a statistically guided multi-colour space stretch (SGMCSS) module. This module randomly selects one of the contrast, saturation, or gamma-corrected versions of the input image and applies a random colour space stretch to it. The colour space stretch is guided by the statistical properties of the input image, which ensures visual consistency with the raw input image. This process generates a reference map that is used by the U-Net-like conditional variational autoencoder (cVAE) module to extract features for feeding to the probabilistic adaptive instance normalization (PAdaIN) block that encodes feature uncertainties for final enhanced image generation. The randomness introduced in the SGMCSS module helps to create diverse pseudo-enhanced images, which can improve the generalization ability of the model and reduce overfitting.
During training, the cVAE network is designed and trained to minimize the reconstruction loss and the KL divergence loss. The reconstruction loss measures the difference between the input image and the reconstructed image, while the KL divergence loss measures the difference between the learned prior distribution of the latent code and the standard normal distribution.
During testing, the cVAE network is used to sample from the learned prior distribution of the latent code to generate new images.

\subsection{Comparison with Uncertainty-Inspired Underwater Image Enhancement}

This subsection provides a detailed comparison between the proposed UDNet framework and the uncertainty-inspired underwater image enhancement method presented by \citep{fu2022uiu} to highlight the unique contributions of our approach. Both methods utilize uncertainty to enhance underwater images; however, their approach to modeling, leveraging, and learning from uncertainty is markedly different.

\subsubsection*{Similarities}

\begin{itemize}
    \item Both UDNet and \citep{fu2022uiu}'s method address the challenge of underwater image enhancement by incorporating \textbf{uncertainty modeling}. Both approaches recognize that the enhancement process has inherent uncertainty due to factors such as varied lighting conditions, water turbidity, and the presence of diverse underwater environments.
\end{itemize}

\subsubsection*{Differences}

\begin{itemize}
    \item \textbf{Uncertainty Modeling}: UDNet employs a framework that models \textit{distributional uncertainty} through adaptive reference selection and probabilistic feature normalization. This approach considers the uncertainty inherent in the potential distributions of enhanced images. In contrast, \citep{fu2022uiu}'s method focuses on modeling \textit{pixel-level uncertainty}, using learned uncertainty maps to refine the enhanced images and quantify uncertainty on a per-pixel basis.
    
    \item \textbf{Reference Map Generation}: UDNet uses a \textit{Statistically Guided Multi-Colour Space Stretch (SGMCSS) module} to generate its reference maps. This module adaptively adjusts the contrast and saturation of the input image and applies gamma correction to generate diverse reference images that are statistically consistent with the input. In contrast, \citep{fu2022uiu}'s method relies on existing UIE algorithms to generate a set of potential reference images.

    \item \textbf{Learning Approach}: UDNet uses a fully \textit{unsupervised approach}, learning from randomly selected, statistically guided reference images. This means it does not require manually annotated or paired training data. \citep{fu2022uiu}, on the other hand, employ a \textit{supervised learning approach}, which requires a manually created set of potential reference images for training.
\end{itemize}

\subsection{Reference Maps Generation}\label{secref}
The main challenge when training deep learning networks for underwater image enhancement is the limited availability of reference maps (labels) for degraded input images.
To address this issue, we auto-generated reference maps based on Underwater Image Enhancement Benchmark Dataset (UIEBD) \citep{Li2020AnBeyond}, which contains real-world underwater images and corresponding reference maps generated using 12 state-of-the-art enhancement algorithms. 
\subsubsection{Autogenration of three reference maps from the input image}
In the original UIEBD, volunteers were asked to compare the enhanced results and subjectively select the best one as the final reference image. However, our reference map generation process uses the same intuition without human intervention. 
Using the degraded input image (original image), as shown in the first step of \cref{fig:8}, we generate three enhanced reference maps by three enhancement algorithms, one of which is randomly selected to introduce uncertainty into our training dataset.
"randomly selected" here refers to the three enhanced reference maps generated by three enhancement algorithms using the degraded input image. One of these three maps is randomly selected to introduce uncertainty into the training dataset.
It is worth mentioning that adding more enhanced reference maps did not increase the model accuracy as discussed in more detail in \cref{secablation}.

The three methods that we chose to introduce uncertainty into the dataset were contrast and saturation adjustment, as well as gamma correction on the original images. These methods were chosen because they can effectively simulate the distortions commonly found in underwater images, such as changes in contrast, saturation, brightness, and colours. As shown in \cref{equ3}, the contrast and saturation adjustment was performed using a linear transformation formula, where the adjustment coefficient $\alpha$ was the same for all pixels for contrast adjustment and varied for each pixel for saturation adjustment.

\begin{equation}\label{equ3}
    y = (x - m) \times \alpha + x, 
\end{equation}
where $x$ and $y$ refer to the degraded and enhanced images, respectively, $m$ denotes the mean of each channel, and $\alpha$ is the adjustment coefficient.

In \cref{equ3}, the adjustment coefficient $\alpha$ plays distinct roles in contrast and saturation adjustments. For contrast adjustment, $\alpha$ remains constant for all pixels, ensuring a uniform contrast enhancement across the entire image. Conversely, for saturation adjustment, $\alpha$ varies for each pixel, enabling fine-grained and localized adjustments to saturation levels. This pixel-wise variation allows for targeted enhancement of specific image areas without influencing others. This difference in $\alpha$ application is crucial, as it results in either global contrast enhancement or localized saturation adjustments, both contributing to a more precise and controlled image enhancement process.

Our approach has several advantages, including saving time and increasing reliability compared to using human observers to generate reference maps. We evaluated the effectiveness of the generated reference maps in \cref{quntres} and \cref{qultres} by comparing the enhanced results to the subjective selections made by volunteers in the original UIEBD dataset.
Our goal was to create uncertain labels that would reflect the uncertainty in the ground truth recording, rather than significantly altering the original labels. To achieve this, we utilized a Statistically Guided Multi-Colour Space Stretch (SGMCSS) or Colour Correction module.

%\subsection{Colour Correction Module}
\subsubsection{Statistically Guided Multi-Colour Space Stretch for Colour Correction}

To improve the visual quality of the reference images used as pseudo-labels, a multi-scale statistically guided multi-colour space stretch module is developed. The term 'multi-scale' refers to the different levels of abstraction in the feature extraction process. The goal of this module is to improve the colour and contrast of the randomly chosen reference maps, which guide the network's unsupervised learning process.

This is obtained by transforming the reference map Red Green Blue (RGB) values to the optimal RGB values, which involves determining the proper camera white-balance for colour-neutral subjects, as well as removing the effects of lens flare and red-green chromatic aberration. This could be useful when dealing with oversaturated images.
The SGMCSS is designed for the case where the mean and standard deviation of the red green and blue colour values are known. This module uses a non-parametric approach to colour correction \citep{Xiao2022usln}, which is able to accommodate new statistical distributions of the pixel values in the red, green and blue colour channels. 
The SGMCSS consists of  two main components: a dual-statistic balance module and a multi-colour space stretch module. 

\textbf{In the dual-statistic  balance module}, the image is processed by two different modules that use statistics of the image (average and maximum values) to correct its colour balance. 
The output is then enhanced using two residual-enhancement modules to recover lost details. 

The first residual-enhancement module is based on Grey World (GW) theory.
The Gray World theory is a method for colour correction in images. It is based on the assumption that the average colour of objects in a perfect image is grey, which means that the average values of the R, G, and B channels are equal. This means that the scale factors for each channel, $e_{R}$, $e_{G}$, and $e_{B}$, can be determined using the GW theory:
\begin{equation}\label{e5}
\setlength\abovedisplayskip{3pt}%shrink space
\setlength\belowdisplayskip{3pt}
x^{GW} = Conv_{1 \times 1}(x) \circ \overline{A},
\end{equation}
where $\overline{A}=[\frac{1}{A_{R}},\frac{1}{A_{G}},\frac{1}{A_{B}}] \in \mathbb{R}^{3 \times 1}$, $A_{c}$ denotes the average value of $c$ channel in the original image, and $\circ$ denotes pixel-wise multiplication.

 % was blue
A $1 \times 1$ convolution operation (\(\operatorname{Con}_{1 \times 1}(x)\)) is used to reduce the number of channels in the input image or to combine information from different channels. In this case, it is used to adjust the contrast, saturation, and gamma correction of the raw underwater image. and then multiplied element-wise (\(\circ\)) with a matrix \(\bar{A}\) to obtain the output image (\(x^{G W}\)).

The second residual-enhancement module is based on the White Patch (WP) algorithm.
The White Patch algorithm is another method for colour correction in images. It is based on the assumption that the maximum response of the RGB channels in an image is caused by a white patch in the scene. This white patch is assumed to reflect the colour of the light in the scene, so the largest value in the RGB channels is used as the source of light. Based on this hypothesis, the scale factors for each channel can be expressed as:
\begin{equation}\label{e8}
\setlength\abovedisplayskip{3pt}%shrink space
\setlength\belowdisplayskip{3pt}
x^{WP} = Conv_{1 \times 1}(x) \circ \overline{M},
\end{equation}
where $\overline{M}=[\frac{1}{M_{R}},\frac{1}{M_{G}},\frac{1}{M_{B}}] \in \mathbb{R}^{3 \times 1}$, $M_{c}$ denotes the maximum value of $c$ channel in original image.

The two residual-enhancement results are merged and passed to the stretch module as follows:
\begin{equation}\label{e9}
\setlength\abovedisplayskip{3pt}%shrink space
\setlength\belowdisplayskip{3pt}
x^{DSB} = Conv_{3 \times 3}(x^{GW}) + Conv_{3 \times 3}(x^{WP}),
\end{equation}
where $x^{DSB}$ represents the  result enhanced by the dual-statistic  balance module.

\textbf{In the multi-colour space stretch module}, the image is transformed into different colour spaces (HSI and Lab) and processed by a trainable module to improve contrast. The original image is also enhanced and added to the stretched version as follows:
\begin{equation}\label{e13}
\setlength\abovedisplayskip{3pt}%shrink space
\setlength\belowdisplayskip{3pt}
x^{final} = Conv_{3 \times 3}(x^{r}) + Conv_{3 \times 3}(x^{h}) + Conv_{3 \times 3}(x^{l}),
\end{equation}
Where $x^{r}$, $x^{h}$, $x^{l}$ denote the histogram stretched pixel value in RGB, HSI, and Lab colour spaces, respectively.

In the RGB colour space, the red, green, and blue channels are individually stretched based on their statistical properties. In the HSI colour space, only the saturation (S) and intensity (I) channels are stretched, while the hue (H) channel is preserved. For the Lab colour space, the a and b channels, representing colour-opponent dimensions, are stretched, while the L channel, representing lightness, is maintained.

The output is then converted back to the RGB colour space and merged together by going through $3 \times 3$ convolutional layer and pixel-wise add up. 
Overall, this technique can improve the visual quality of the reference map that will be passed to the next building block of UDNet, i.e. the feature extractor module (see \cref{fig:8}), by correcting colour balance and enhancing contrast.

\subsection{Feature Extraction} \label{sec:featExt}
The next abstract building block of UDNet, as shown in \cref{fig:8}, is its feature extractor block. UDnet uses a two-branch U-Net-based feature extractor to map the input images to representations. These representations are then fed into the PAdaIN module, which transforms the enhancement statistics of the input to create the enhanced image as explained in \ref{padin}.

The training branch of the feature extractor is used to construct posterior distributions using the raw original underwater image and its corresponding reference map image as inputs. The test branch, on the other hand, is used to estimate the prior distribution of a single raw underwater image.

The PAdaIN block is used to encode the uncertainty in the input image, allowing UDNet to generate multiple enhanced versions of the image that capture the different possible interpretations of the original image.
To achieve this, UDnet uses a prior/posterior block to build the distribution of possible enhancements. This block is designed to construct both a mean and a standard deviation distribution, using ${1 \times 1}$ convolutions to transform the input data matrix into a series of distributions that capture the uncertainty in the input image. 

In the training stage, the input image and its corresponding reference image are used to learn the \texttt{posterior}  distributions of the latent codes as follows: 
\begin{equation}\label{equ4}
{\bm{a}} \sim  \mathcal{N}_{\rm{mean}}\left( {\bm{\mu} \left( {{\bm{y}},{\bf{x}}} \right),\bm{\sigma}^2 \left( {{\bm{y}},{\bf{x}}} \right)} \right),
\end{equation}
\vspace{-4mm}
\begin{equation}\label{equ5}
{\bm{b}} \sim  \mathcal{N}_{\rm{std}}\left( {\bm{m} \left( {{\bm{y}},{\bf{x}}} \right),\bm{v}^2 \left( {{\bm{y}},{\bf{x}}} \right)} \right),
\end{equation}
where $\bm{a}$ and ${\bm{b}}$ are two random samples from the mean and standard deviation posterior distributions, $\mathcal{N}_{\rm{mean}}$ and $\mathcal{N}_{\rm{std}}$ are the $N$-dimensional Gaussian distribution of the mean and standard deviation, and ${\bm{y}}$ and ${\bm{x}}$ are the reference image and the raw input image, respectively.

\cref{equ4} represents the distribution of the feature activations ($a$) in the UDnet's conditional variational autoencoder (cVAE) module. The distribution is assumed to be normal (represented by the symbol \(\mathcal{N}\)) with mean \(\mu(\boldsymbol{y}, \mathbf{x})\) and variance \(\sigma^{2}(\boldsymbol{y}, \mathbf{x})\). The mean and variance are functions of the input image (\(\mathbf{x}\)) and the reference map (\(\boldsymbol{y}\)) generated by the statistically guided multi-colour space stretch (SGMCSS) module.

\cref{equ5}  represents the distribution of the feature uncertainties (represented by the vector \(\boldsymbol{b}\)) in the UDnet's probabilistic adaptive instance normalization (PAdaIN) block. The distribution is also assumed to be normal, but with a different subscript \(\mathrm{s}\) to distinguish it from the distribution in the cVAE module. The mean and variance of the distribution are functions of the input image and the reference map, represented by \(\boldsymbol{m}(\boldsymbol{y}, \mathbf{x})\) and \(\boldsymbol{v}^{2}(\boldsymbol{y}, \mathbf{x})\), respectively.

Overall, these equations describe the probabilistic nature of UDnet, which allows it to model the uncertainty in the input data and generate enhanced images that are consistent with the input image and reference map. The use of probabilistic distributions also enables UDnet to learn from a limited amount of data without the need for manual human annotation.

Once these distributions have been constructed, random samples are extracted from them and injected into the PAdaIN module, where they are used to transform the statistics of the received features.

In the testing stage, the latent codes generated for PAdaIN are determined only by the input image to learn the \texttt{prior} distributions of the latent codes as follows:
\begin{equation}\label{equ6}
 {\bm{a}} \sim \mathcal{N}_{\rm{mean}}\left( {\bm{\mu}\left( {{\bf{x}}} \right),\bm{\sigma}^2\left( {{\bf{x}}} \right)} \right),
\end{equation}
\vspace{-4mm}
\begin{equation}\label{equ7}
{\bm{b}} \sim  \mathcal{N}_{\rm{std}}\left( {\bm{m}\left( {{\bf{x}}} \right),\bm{v}^2\left( {{\bf{x}}} \right)} \right),
\end{equation}
where $\bm{a}$ and ${\bm{b}}$ are two random samples from the mean and standard deviation prior distributions, $\mathcal{N}_{\rm{mean}}$ and $\mathcal{N}_{\rm{std}}$ are the $N$-dimensional Gaussian distribution of the mean and standard deviation, respectively and ${\bm{x}}$ is the raw input image.

The UDnet model is applied multiple times to the same input image in order to generate multiple enhancement variants. This is done by re-evaluating only the PAdaIN module and the output block, without retraining the entire model, which makes UDNet very efficient.  
The resulting diverse enhancement samples are then used for Maximum Probability estimation that takes the enhancement sample with the maximum probability as the final estimation.

\subsubsection{Loss Function} \label{seclos}
The training process for UDnet follows the standard procedure for training a cVAE model, which involves minimizing the variational lower bound. However, our approach has an additional step of finding a meaningful embedding of enhancement statistics in the latent space. This is achieved through the use of a posterior network (as shown in \cref{fig:8}), which learns to recognize posterior features and map them to posterior distributions of the mean and standard deviation. Random samples from these distributions can be used to formalize the enhanced results. This approach allows for the incorporation of uncertainty into the enhancement process, which can improve the accuracy and reliability of the resulting images.

During the training process, the PAdaIN module is used to predict the enhanced image by receiving random samples ${\bm{a}}$ and ${\bm{b}}$ from \cref{equ4} and \cref{equ5}, respectively. The enhancement loss (\cref{equ9}) is calculated based on the differences between the predicted image and the reference map, and is used to penalize the model if the output deviates from the reference. %The exact formula for enhancement loss is:
\begin{equation}\label{equ9}
{L_{\rm{e}}} = {L_{\rm{mse}}} + \lambda {L_{\rm{vgg16}}},
\end{equation}
where ${L_{\rm{mse}}}$ denotes the mean square error loss and ${L_{\rm{vgg16}}}$ denotes the perceptual loss \citep{Johnson2016PerceptualSuper-resolution}, $\lambda$ refers to a weight parameter. 

The mean square error loss $L_{\rm{mse}}$ and the perceptual loss $L_{\rm{vgg16}}$ are two common metrics used to evaluate the performance of image enhancement algorithms. The mean square error loss measures the average squared difference between the predicted and reference images, while the perceptual loss, which was introduced by Johnson et al. \citep{Johnson2016PerceptualSuper-resolution}, measures the differences between the high-level features of the predicted and reference images. The weight $\lambda$ is used to control the relative importance of these two loss terms in the overall enhancement loss $L_{\rm{e}}$. For example, if $\lambda$ is set to a high value, the model will be more heavily penalized for large differences between the predicted and reference images, while if $\lambda$ is set to a low value, the model will be less sensitive to such differences. The specific values of $\lambda$ used in the training process will depend on the characteristics of the dataset and the desired performance of the model.

In addition to minimizing the enhancement loss $L_{\rm{e}}$, the training process for UDnet also involves using Kullback-Leibler (KL) divergences $D_{\rm{KL}}$ to align the posterior distributions with the prior distributions (\cref{equ10} and \cref{equ11}). 
\begin{equation}\label{equ10}
L_m = {D_{{\rm{KL}}}}\left( {{\mathcal{N}_{\rm{mean}}}\left( {{\bf{x}}} \right)\left\| {{\mathcal{N}_{\rm{mean}}}\left( {{\bm{y}},{\bf{x}}} \right)} \right.} \right),
\end{equation}
\vspace{-4mm}
\begin{equation}\label{equ11}
L_s = {D_{{\rm{KL}}}}\left( {{\mathcal{N}_{\rm{std}}}\left( {{\bf{x}}} \right)\left\| {{\mathcal{N}_{\rm{std}}}\left( {{\bm{y}},{\bf{x}}} \right)} \right.} \right),
\end{equation}
where $\bm{m}$ and ${\bm{s}}$ are the mean and the standard deviation, respectively.
KL divergence is a measure of the difference between two probability distributions and can be used to compare the posterior distributions learned by the model with the prior distributions that are assumed to represent the distribution of latent variables in the training data. By minimizing the KL divergences between the posterior and prior distributions, the model is able to learn a more accurate representation of the latent space, which can improve the quality of the enhanced images.  %The exact formula for Kullback-Leibler (KL) divergences is:

The total loss function used for training UDnet is the weighted sum of the enhancement loss $L_{\rm{e}}$ and the KL divergences $D_{\rm{KL}}$ between the posterior and prior distributions,
\begin{equation}\label{equ12}
L = {L_e} + \beta (L_m + L_s),
\end{equation}
where $\beta$ is a weight parameter, whose value %The specific weight $\beta$ used for these loss terms will 
depends on the dataset's characteristics and the model's desired performance. By minimizing this total loss function, $L$, the model is able to learn an effective mapping from the input degraded images to the corresponding enhanced images, while also aligning the posterior and prior distributions in the latent space. This allows the model to generate high-quality enhanced images while also incorporating uncertainty into the enhancement process.

\subsection{Probabilistic Adaptive Instance Normalization (PAdaIN)} \label{padin}

The final abstract building block of UDNet, as shown in \cref{fig:8} is PAdaIN block.
The goal of UDnet is to adjust the appearance of underwater images, such as the colours and contrasts, without altering the content of the image. This is important because it allows the enhanced images to be more visually appealing and easier to interpret, without compromising the integrity of the original image. Therefore, We use a probabilistic adaptive instance normalization (PAdaIN) to capture these properties.

The core component of UDNet's probabilistic framework is PAdaIN \citep{fu2022uiu}, a modified version of the AdaIN algorithm specifically designed for underwater image enhancement. PAdaIN leverages the uncertainty distribution learned during training to encode feature uncertainties, allowing it to generate multiple enhanced versions of the input image. These multiple versions reflect the inherent ambiguity in underwater image enhancement, as there is no single correct enhancement for a given image. By generating a distribution of possible enhancements, UDNet provides a more comprehensive and flexible approach to underwater image enhancement.

Unlike \citep{fu2022uiu}, which utilizes static uncertainty maps to guide image enhancement, our method employs adaptive uncertainty modelling by randomly selecting reference images during training. This approach encourages the network to learn from a diverse set of potential image distributions, resulting in a more robust and flexible enhancement framework. Additionally, our PAdaIN layer further refines feature alignment by accounting for variability in the latent space, enabling our model to generalize effectively to datasets with diverse characteristics. These advancements collectively extend \citep{fu2022uiu} method by addressing its limitations in handling diverse underwater environments and ensuring consistent enhancement quality.

However, AdaIN relies on the availability of known content and style images, which is not always the case in underwater image enhancement processes. To address this issue, PAdaIN introduces random samples from the posterior distributions of the mean and standard deviation as the parameters of the AdaIN operation, which can be formulated as: \begin{equation}\label{equ2}
{\rm{PAdaIN}} \left( {\bf{x}} \right) = {\bm{b}}\left( {\frac{{{\bf{x}} - \bm{\mu}\left( {\bf{x}} \right)}}{{\bm{\sigma}\left( {\bf{x}} \right)}}} \right) + {\bm{a}},
\end{equation}
where $\bm{b}$ and $\bm{a}$ are two random samples from the posterior distributions of the mean and standard deviation, respectively.

These posterior distributions are learned using a cVAE, which was described in \ref{sec:featExt}. This allows PAdaIN to generalize the AdaIN algorithm and apply it to underwater image enhancement without the need for known content and style images. Overall, PAdaIN is able to capture the important appearance-related features of the input image and use them to generate enhanced images that maintain the integrity of the original image.

It is worth noting that in contrast to other approaches that consider the variance of the image, such as GAN, PAdaIN is based on the statistical distribution of the image features, which are invariant to transformations like colour transformation. This is done by conditioning the network on training images and their reference map, which, along with the use of a differentiable approximation of the uncertainty, make UDnet easily trainable with a single backward pass.

%%%%%%%%%%%%%%%%%%%%%%%%%%%%%%%%%%%%%%%%%%%%%%%%%%%%%%%%%%%%%%%%
\section{Experiments} \label{secresult}
In this section, we perform several experiments to evaluate the performance of our proposed method. We will first describe the utilized datasets, evaluation
metrics and implementation details. 
Then, we quantitatively and qualitatively evaluate our  model against 10 popular image enhancement models on 8 public datasets. 
Finally, we will demonstrate the significance of our work through a visual perception improvement test.

\subsection{Datasets}
We used eight publicly available datasets for our model's performance verification. These datasets are: EUVP \citep{Islam2020FastPerception}, UFO \citep{Islam2022seu}, UIEBD \citep{Li2020AnBeyond}, DeepFish  \citep{Saleh2020}, FISHTRAC \citep{Mandel2023DetectionDetections}, FishID \citep{lopiz2021eco}, RUIE \citep{Liu2020Real-worldLight}, SUIM \citep{Islam2020}. 
Details of these datasets can be found in   \cref{table3}.
In EUVP \citep{Islam2020FastPerception}, UFO \citep{Islam2022seu}, and UIEBD \citep{Li2020AnBeyond}, there are many paired images and unpaired images which were divided as shown in \cref{table3}. The paired images are the ones that have ground truth. 
The rest of the datasets have only unpaired images.
In our experiment, we used only UIEBD \citep{Li2020AnBeyond} for training in an unsupervised way without the ground truth. We used the other datasets for performance evaluation.

\begin{table}[h!]%[ht] 
\centering
\centering
\caption{THE DATASETS USED IN OUR RESEARCH. THE NUMBERS REPRESENT THE AMOUNT OF IMAGES IN SETS}
\label{table3}
\small
\setlength{\tabcolsep}{8pt}
\begin{spacing}{1.39}
\begin{tabular}{l|cc|cccc}
\hline
\multirow{2}{*}{Datasets} & \multicolumn{2}{c|}{Train} & \multicolumn{2}{c}{Test} \\
\cline{2-5}
             &Paired    & Unpaired  & Paired    & Unpaired      \\
\hline
EUVP \citep{Islam2020FastPerception}    & 3700  & 3140   & 515     & -    \\
\hline
UFO \citep{Islam2022seu}     & 1500  & -   & 120     & -    \\
\hline
UIEBD \citep{Li2020AnBeyond}     & 800 & -    & 90      & 60       \\
\hline
DeepFish  \citep{Saleh2020}  & -    & 3200    & -      & 600       \\
\hline
FISHTRAC \citep{Mandel2023DetectionDetections}    & -  & 600    & -      & 71       \\
\hline
FishID \citep{lopiz2021eco}    & -  & 7093    & -      & 6897       \\
\hline
RUIE \citep{Liu2020Real-worldLight}    & -  & 2904    & -      & 726       \\
\hline
SUIM \citep{Islam2020}    & -  & 1525    & -      & 110       \\
\hline
\end{tabular}
\end{spacing}
% \label{tab2}
\vspace{-1em}
\end{table}

\subsection{Evaluation Metrics}
Evaluation metrics for image enhancement are often based on natural image statistics. Perceptual and structural image qualities can be judged in different ways. We employ four full-reference evaluation metrics and three no-reference evaluation metrics for evaluating the quantitative performance of our image enhancement model.
Specifically, 1) The full-reference evaluation metrics consist of Peak Signal-to-Noise Ratio (PSNR) \citep{Wang2004ImageSimilarity}, Structural Similarity (SSIM) \citep{Wang2004ImageSimilarity}, Most Apparent Distortion (MAD) \citep{mad2010},  and Gradient Magnitude Similarity Deviation (GMSD)  \citep{Xuegmsd}, which are used for paired test sets (EUVP \citep{Islam2020FastPerception}, UFO \citep{Islam2022seu}, UIEBD \citep{Li2020AnBeyond}). 
A higher PSNR or a lower MAD score means that the output image and the label image are closer in perceptual content, while a higher SSIM or a lower GMSD score means that the two images are more structurally similar.
2) The no-reference evaluation metrics are: Underwater Image Quality Measure (UIQM) \citep{Panetta2016}, Multi-scale Image Quality Transformer (MUSIQ) \citep{MUSIQ2021}, and Natural Image Quality Evaluator (NIQE) \citep{NIQE2013} which are used for unpaired test sets (DeepFish  \citep{Saleh2020}, FISHTRAC \citep{Mandel2023DetectionDetections}, FishID \citep{lopiz2021eco}, RUIE \citep{Liu2020Real-worldLight}, SUIM \citep{Islam2020}). The UIQM is the linear combination of three underwater image attribute measures: the underwater image colourfulness measure (UICM), the underwater image sharpness measure (UISM), and the underwater image contrast measure (UIConM).
A higher UIQM and MUSIQ or a lower NIQE score suggests a better human visual perception.
However, it is worth noting that these no-reference metrics cannot accurately reflect the quality of an image in some cases, so scores of UIQM, MUSIQ, and NIQE are only provided as references for our study. We will present enhanced unpaired images in the visual comparisons section for readers to assess.

We chose these metrics because they collectively provide a comprehensive evaluation of our image enhancement model. The full-reference metrics allow us to compare the enhanced images directly with the original images, providing a measure of the fidelity and structural similarity of our enhancements. The no-reference metrics offer an assessment of the image quality from a human visual perception perspective, which is crucial as the ultimate goal of our work is to improve the visual quality of underwater images for human viewers. These metrics together ensure a robust and thorough evaluation of our method. For example, PSNR measures the fidelity of the enhanced image to the original image, while SSIM takes into account the structural information in the image. MAD measures the most apparent distortion between the enhanced and original images, while GMSD measures the similarity of gradient magnitude between the two images.   While we acknowledge that no single metric can fully capture the quality of an enhanced image, we believe that our choice of these seven metrics provides a meaningful and comprehensive evaluation of our proposed method.

\begin{sidewaystable}[!ph]  
% \begin{table*}[ht]
\centering
\caption{COMPARISON AGAINST PUBLISHED WORKS ON THREE \textit{PAIRED} DATASETS (EUVP \citep{Islam2020FastPerception}, UFO \citep{Islam2022seu} AND UIEBD \citep{Li2020AnBeyond}).\\ UNDERWATER IMAGE ENHANCEMENT PERFORMANCE METRIC IN TERMS OF AVERAGE PSNR \citep{Wang2004ImageSimilarity}, SSIM \citep{Wang2004ImageSimilarity}, MAD \citep{mad2010} AND GMSD  \citep{Xuegmsd} VALUES ARE SHOWN, WHERE ($\uparrow$) MEANS HIGHER IS BETTER AND ($\downarrow$) MEANS LOWER IS BETTER. WE REPRESENT THE BEST TWO RESULTS IN \textcolor{red}{RED} AND \textcolor[rgb]{0,0.69,0.941}{BLUE} COLOURS. \\$^{*}$ THE MODEL TRAINED ON UIEBD \citep{Li2020AnBeyond} DATASET WITH LABEL.}
\label{table1}
\small
\begin{spacing}{2.1}
\setlength{\tabcolsep}{2pt}
\begin{tabular}{l|cccc|cccc|cccc} 
\hline
\multirow{2}{*}{Method} & \multicolumn{4}{c|}{EUVP} & \multicolumn{4}{c|}{UFO} & \multicolumn{4}{c}{UIEBD}\\
\cline{2-13}
 &PSNR $\uparrow$ & SSIM $\uparrow$      & MAD $\downarrow$    & GMSD $\downarrow$ 
 &PSNR $\uparrow$ & SSIM $\uparrow$      & MAD $\downarrow$    & GMSD $\downarrow$ 
 &PSNR $\uparrow$ & SSIM $\uparrow$      & MAD $\downarrow$    & GMSD $\downarrow$\\
 % PSNR $\uparrow$       & SSIM $\uparrow$  &MSE $\downarrow$   & PSNR $\uparrow$       & SSIM $\uparrow$ & PSNR $\uparrow$       & SSIM $\uparrow$ & SSIM $\uparrow$\\
\hline
 {CLAHE} \citep{CLAHE} & 18.97 & 0.726 & 138.6 & 0.090 & 18.76 & 0.701 & 143.7 & 0.098 & 20.64 & 0.821 & 100.9 & 0.053 \\ 
\hline
 {IBLA} \citep{IBLA2017} & \textcolor[rgb]{0,0.69,0.941}{22.62} & 0.719 & 97.78 & 0.068 & 20.71 & 0.671 & 122.6 & 0.082 & 17.56 & 0.614 & 141.2 & 0.126 \\ 
\hline
 {PIFM$^{*}$} \citep{PIFM2021} & 20.17 & 0.747 & 113.5 & 0.0719 & 20.63 & 0.728 & 118.2 & 0.076 & 23.62 & 0.852 & 80.80 & 0.056 \\ 
\hline
 {PUIEnet$^{*}$} \citep{fu2022uiu} & 21.01 & 0.770 & \textcolor[rgb]{0,0.69,0.941}{94.55} & 0.052 & 21.38 & 0.737 & \textcolor[rgb]{0,0.69,0.941}{102.3} & \textcolor[rgb]{0,0.69,0.941}{0.057} & 23.74 & 0.844 & 79.36 & 0.057 \\ 
\hline
 {RGHS} \citep{Huang2018Shallow-waterAcquisition} & 21.13 & 0.753 & 98.40 & 0.056 & 20.74 & 0.730 & 112.8 & 0.066 & 23.57 & 0.803 & 81.02 &  0.053 \\ 
\hline
 {UCM} \citep{Iqbal2010EnhancingMethod} & 20.91 & 0.767 & 99.37 & 0.062 & 20.34 & \textcolor[rgb]{0,0.69,0.941}{0.743} & 110.0 & 0.068 & 22.03 & 0.815 & 92.95 & 0.067 \\ 
\hline
 {UDCP} \citep{Drews2013TransmissionImages} & 15.80 & 0.572 & 136.8 & 0.098 & 15.95 & 0.561 & 148.1 & 0.111 & 13.47 & 0.548 & 139.0 & 0.118 \\ 
\hline
 {ULAP} \citep{Song2018ARestoration} & 21.91 & 0.730 & 108.4 & 0.071 & \textcolor{red}{21.98} & 0.729 & 116.7 & 0.071 & 18.95 & 0.718 & 113.0 & 0.085 \\ 
\hline
 {USLN$^{*}$} \citep{Xiao2022usln} & 20.87 & \textcolor[rgb]{0,0.69,0.941}{0.771} & 94.62 & \textcolor[rgb]{0,0.69,0.941}{0.050} & 20.73 & \textcolor{red}{0.749} & 105.0 & 0.057 & \textcolor[rgb]{0,0.69,0.941}{24.04} & \textcolor[rgb]{0,0.69,0.941}{0.849} & 78.91 & 0.057 \\ 
\hline
 {Wavenet$^{*}$} \citep{Wavelength} & 20.25 & 0.753 & 109.3 & 0.067 & 20.98 & 0.736 & 115.1 & 0.071 & \textcolor{red}{24.61} & \textcolor{red}{0.881} & \textcolor{red}{68.08} & \textcolor[rgb]{0,0.69,0.941}{0.045} \\ 
 
\hline
 {UDnet (ours)} & \textcolor{red}{22.96} & \textcolor{red}{0.771} & \textcolor{red}{87.67} & \textcolor{red}{0.049} & \textcolor[rgb]{0,0.69,0.941}{22.43} & 0.738 & \textcolor{red}{99.53} & \textcolor{red}{0.053} & 22.23 & 0.812 & \textcolor[rgb]{0,0.69,0.941}{74.21} & \textcolor{red}{0.043} \\
\hline
\end{tabular}
\end{spacing}
% \label{tab1}
% \vspace{-1em}
% \end{table*}
\end{sidewaystable}

\begin{sidewaystable}[!ph]
\centering
\caption{COMPARISON AGAINST PUBLISHED WORKS ON FIVE \textit{UNPAIRED} DATASETS (DeepFish \citep{Saleh2020}, FISHTRAC \citep{Mandel2023DetectionDetections},  FishID \citep{lopiz2021eco}, RUIE \citep{Liu2020Real-worldLight}, AND SUIM \citep{Islam2020}).\\ UNDERWATER IMAGE ENHANCEMENT PERFORMANCE METRIC IN TERMS OF AVERAGE UIQM \citep{Panetta2016}, MUSIQ \citep{MUSIQ2021} AND NIQE \citep{NIQE2013} VALUES  ARE SHOWN, WHERE ($\uparrow$) MEANS HIGHER IS BETTER, AND ($\downarrow$) MEANS LOWER IS BETTER. WE REPRESENT THE BEST TWO RESULTS IN \textcolor{red}{RED} AND \textcolor[rgb]{0,0.69,0.941}{BLUE} COLOURS.} 
% $^{*}$ THE MODEL TRAINED ON UIEBD DATASET WITH LABEL.}
\label{table2}
\small
\begin{spacing}{2.1}
\setlength{\tabcolsep}{2pt}
\begin{tabular}{l|ccc|ccc|ccc|ccc|ccc} 
\hline
\multirow{2}{*}{Method} & \multicolumn{3}{c|}{DeepFish} & \multicolumn{3}{c|}{FISHTRAC} & \multicolumn{3}{c|}{FishID} & \multicolumn{3}{c|}{RUIE} & \multicolumn{3}{c}{SUIM}\\
\cline{2-16}
 &UIQM$\uparrow$ & MUSIQ$\uparrow$      & NIQE$\downarrow$    & UIQM$\uparrow$ 
 &MUSIQ$\uparrow$ & NIQE$\downarrow$      & UIQM$\uparrow$    & MUSIQ$\uparrow$ 
 &NIQE$\downarrow$ & UIQM$\uparrow$      & MUSIQ$\uparrow$    & NIQE$\downarrow$
 &UIQM$\uparrow$ & MUSIQ$\uparrow$      & NIQE$\downarrow$    \\
 % PSNR $\uparrow$       & SSIM $\uparrow$  &MSE $\downarrow$   & PSNR $\uparrow$       & SSIM $\uparrow$ & PSNR $\uparrow$       & SSIM $\uparrow$ & SSIM $\uparrow$\\
\hline
 {CLAHE} \citep{CLAHE} & 3.136 & 25.67 & 4.10 & 2.686 & 48.94 & 3.78 & \textcolor{red}{2.631} & 37.01 & 5.20 & 3.028 & \textcolor{red}{34.98} & 4.48 & \textcolor{red}{2.914} & 58.40 & \textcolor[rgb]{0,0.69,0.941}{3.66} \\ 
\hline
 {IBLA} \citep{IBLA2017} & 1.993 & \textcolor{red}{43.34} & 6.31 & 1.704 & 55.36 & 6.20 & 1.745 & 44.45 & 6.69 & 2.577 & 32.60 & 4.68 & 1.839 & 58.17 & 4.10 \\ 
\hline
 {PIFM} \citep{PIFM2021} & 3.275 & 27.39 & 4.29 & 2.892 & 48.38 & 4.17 & 2.424 & 41.14 & \textcolor[rgb]{0,0.69,0.941}{5.12} & 3.087 & \textcolor[rgb]{0,0.69,0.941}{33.29} & 4.51 & 2.694 & 58.56 & 3.82 \\ 
\hline
 {PUIEnet} \citep{fu2022uiu} & 3.209 & 28.82 & 4.10 & \textcolor{red}{3.421} & 48.60 & 4.22 & 2.301 & 40.08 & 5.32 & 3.102 & 28.32 & 4.56 & 2.838 & 60.03 & 3.75 \\ 
\hline
 {RGHS} \citep{Huang2018Shallow-waterAcquisition} & 3.150 & \textcolor[rgb]{0,0.69,0.941}{42.26} & 6.72 & 1.913 & \textcolor{red}{57.03} & 5.57 & 1.823 & 44.46 & 5.95 & 2.991 & 30.49 & 4.32 & 2.317 & 59.55 & 3.75 \\ 
\hline
 {UCM} \citep{Iqbal2010EnhancingMethod} & 2.918 & 41.37 & 6.45 & 2.575 & 54.39 & 11.25 & \textcolor[rgb]{0,0.69,0.941}{2.452} & 44.32 & 6.06 & \textcolor[rgb]{0,0.69,0.941}{3.107} & 32.59 & \textcolor[rgb]{0,0.69,0.941}{4.19} & 2.804 & 58.51 & 3.90 \\ 
\hline
 {UDCP} \citep{Drews2013TransmissionImages} & 2.391 & 42.42 & 5.72 & 1.622 & 51.01 & 6.09 & 1.320 & 37.97 & 6.48 & 2.159 & 29.79 & 4.71 & 1.731 & 56.14 & 4.09 \\ 
\hline
 {ULAP} \citep{Song2018ARestoration} & 2.814 & 40.46 & 6.16 & 1.763 & 54.37 & 6.18 & 2.176 & \textcolor[rgb]{0,0.69,0.941}{45.12} & 5.98 & 2.396 & 33.00 & 4.72 & 2.232 & 58.36 & 3.97 \\ 
\hline
 {USLN} \citep{Xiao2022usln} & 3.015 & 32.12 & 4.24 & 3.316 & 49.97 & 4.30 & 2.067 & 44.58 & 6.32 & 3.068 & 32.78 & 4.44 & 2.682 & \textcolor{red}{61.30} & 4.03 \\ 
\hline
 {Wavenet} \citep{Wavelength} & \textcolor{red}{3.304} & 40.14 & \textcolor{red}{3.00} & 3.071 & \textcolor[rgb]{0,0.69,0.941}{56.79} & \textcolor[rgb]{0,0.69,0.941}{3.70} & 2.252 & \textcolor{red}{46.66} & 5.15 & 3.081 & 30.79 & 4.56 & 2.773 & \textcolor[rgb]{0,0.69,0.941}{61.03} & 3.67 \\ 
\hline
{UDnet(ours)} & \textcolor[rgb]{0,0.69,0.941}{3.292} & 24.56 & \textcolor[rgb]{0,0.69,0.941}{4.02} & \textcolor[rgb]{0,0.69,0.941}{3.390} & 50.24 & \textcolor{red}{3.41} & 2.303 & 37.85 & \textcolor{red}{5.11} & \textcolor{red}{3.154} & 26.74 & \textcolor{red}{4.18} & \textcolor[rgb]{0,0.69,0.941}{2.875} & 57.84 & \textcolor{red}{3.65} \\
\hline
\end{tabular}
\end{spacing}
% \label{tab2}
%\vspace{-1em}
\end{sidewaystable}

\begin{sidewaystable}[!ph]
\centering
\caption{COMPARISON AGAINST PUBLISHED WORKS ON TWO \textit{PAIRED} DATASETS (UIEBD \citep{Li2020AnBeyond},  ANDEUVP \citep{Islam2020FastPerception} ) AND TWO \textit{UNPAIRED} DATASETS(UCCS \citep{Liu2020Real-worldLight}, AND UIQS \citep{Liu2020Real-worldLight}) .\\ UNDERWATER IMAGE ENHANCEMENT PERFORMANCE METRIC IN TERMS OF AVERAGE PSNR \citep{Wang2004ImageSimilarity}, SSIM \citep{Wang2004ImageSimilarity},  UIQM \citep{Panetta2016}, UCIQE \citep{Panetta2016},  HIGHER VALUES IS BETTER. WE REPRESENT THE BEST TWO RESULTS IN \textcolor{red}{RED} AND \textcolor[rgb]{0,0.69,0.941}{BLUE} COLOURS.}
\label{table5}
\small
\begin{spacing}{2.1}
\setlength{\tabcolsep}{2.5pt}
\begin{tabular}{l|cccc|cccc|cc|cc} 
\hline
\multirow{2}{*}{Method} & \multicolumn{4}{c|}{UIEBD} & \multicolumn{4}{c|}{EUVP} & \multicolumn{2}{c|}{UCCS} & \multicolumn{2}{c}{UIQS} \\
\cline{2-13}
 &PSNR   & SSIM        & UIQM      & UCIQE   
 &PSNR   & SSIM        & UIQM      & UCIQE   
 &UIQM   & UCIQE        & UIQM      & UCIQE  \\

\hline
 {FIRUA} \citep{Yu2023AnAutoencoder} & \textcolor{red}{27.80} & \textcolor[rgb]{0,0.69,0.941}{0.849} & 3.452 & 0.534 & 20.65 & 0.728 & 2.985 & 0.316 & 3.650 & 0.652 & 3.223 & 0.809 \\ 
\hline
 {RCA-CycleGAN} \citep{Wang2023RCA-CycleGAN:CycleGAN} & 21.28 & 0.804 & 2.987 & 0.280 & {21.32} & \textcolor[rgb]{0,0.69,0.941}{0.829} & 2.834 & 0.337 & 2.942 & 0.671 & 2.975 & 0.832 \\ 
\hline
 {IEFD} \citep{Liu2022UnsupervisedDisentanglement} & 22.01 & 0.793 & 3.234 & 0.317 & 21.17 & 0.705 & {3.148} & {0.363} & 3.199 & 0.598 & 3.379 & 0.827 \\ 
\hline
 {RFHP} \citep{Fu2022UnsupervisedPerspective} & 20.31 & 0.841 & 3.189 & 0.598 & 20.91 & 0.649 & 2.932 & 0.299 & 3.345 & 0.627 & \textcolor{red}{3.788} & 0.801 \\ 
\hline
 {Two-step DA} \citep{Jiang2022Two-stepEnhancementb} & 21.78 & 0.802 & 3.402 & 0.624 & 20.37 & 0.711 & 3.105 & 0.208 & 3.724 & 0.674 & \textcolor[rgb]{0,0.69,0.941}{3.418} & \textcolor[rgb]{0,0.69,0.941}{0.913} \\ 
\hline
 {MCACE} \citep{Zhang2022UnderwaterEnhancement} & 20.69 & 0.784 & 3.293 & 0.587 & 20.89 & 0.698 & 2.927 & 0.309 & 3.933 & 0.587 & 2.851 & 0.856 \\ 
\hline
 {Twin ACL} \citep{Liu2022TwinBeyond} & 22.30 & \textcolor{red}{0.888} & 3.595 & \textcolor[rgb]{0,0.69,0.941}{0.683} & 21.04 & 0.724 & 3.029 & 0.354 & \textcolor[rgb]{0,0.69,0.941}{3.953} & \textcolor[rgb]{0,0.69,0.941}{0.688} & 3.032 & 0.897 \\ 
\hline
 MNIAM \citep{ji2024dual} & 23.11 & 0.833 & 3.590 & 0.607 & 21.49 & 0.752 & \textcolor[rgb]{0,0.69,0.941}{3.251} & 0.619 & 3.719 & 0.653 & 3.111 & 0.848 \\ 
\hline
 LFT-DGAN \citep{zheng2024learnable} & \textcolor[rgb]{0,0.69,0.941}{24.45} & 0.828 & 3.442 & 0.593 & 21.67 & 0.741 & 3.109 & 0.610 & 3.618 & 0.645 & 3.289 & 0.832 \\ 
\hline
 FMTformer \citep{xiang2025fusion} & 22.89 & 0.812 & 3.393 & 0.621 & 20.67 & 0.807 & 3.014 & 0.598 & 3.567 & 0.633 & 3.218 & 0.847 \\ 
\hline
 FDCE-Net \citep{cheng2024fdce} & 23.87 & 0.917 & 3.561 & 0.612 & \textcolor{red}{26.17} & \textcolor{red}{0.893} & 3.214 & 0.645 & 3.833 & 0.681 & 4.253 & 0.598 \\ 
\hline
 Zero-UMSIE \citep{liu2024zero} & 23.45 & 0.841 & \textcolor{red}{4.837} & 0.621 & 22.89 & 0.803 & 3.104 & \textcolor[rgb]{0,0.69,0.941}{0.649} & 3.793 & 0.671 & 3.672 & 0.879 \\ 
\hline
 {UDnet(ours)} & 22.23 & 0.812 & \textcolor[rgb]{0,0.69,0.941}{3.781} & \textcolor{red}{0.745} & \textcolor[rgb]{0,0.69,0.941}{22.96} & {0.771} & \textcolor{red}{3.265} & \textcolor{red}{0.749} & \textcolor{red}{3.974} & \textcolor{red}{0.713} & 3.154 & \textcolor{red}{0.958} \\ 
\hline

\end{tabular}
\end{spacing}
\end{sidewaystable}

\subsection{Implementation Details}
Our models were trained with an input resolution of $256 \times 256$ pixels.
We scale the lowest side of the image to $256$ and then extract random crops of size $256 \times 256$.
We found that for this problem set, a learning rate of $1 \times 10^{-4}$ works the best. It took around $500$ epochs for the model to train on this problem and the batch size was set as 10. Our networks were trained on a  Linux host with a single NVidia GeForce RTX 2080 Ti GPU with $11$ GB of memory, using Pytorch framework.
% \citep{Paszke2019}.  
The training is carried out with ADAM optimizer, and the loss function, as explained in \ref{seclos}, is a combination of the Mean Squared Error (MSE) ${L_{\rm{mse}}}$, the perceptual loss ${L_{\rm{vgg16}}}$ \citep{Johnson2016PerceptualSuper-resolution}, and Kullback-Leibler (KL) divergences ${L_{\rm{kl}}}$.
% \citep{Contreras-Reyes2012}.

In order to boost network generalisation, we augment the training data with rotation, flipping horizontally and vertically. Following \citep{fu2022uiu}, we adopt ${1 \times 1}$ convolutions to broadcast the samples to the desired number of channels before input to PAdaIN with a latent space of a 20-dimensional $N$.

 % was blue
From an implementation and computational point of view, our proposed method, UDnet, is feasible. We implemented UDnet using PyTorch and trained it on a single NVIDIA GeForce RTX 2080 Ti GPU. The training time and the number of parameters for each component of UDnet are as follows: the SGMCSS module has 1.2 million parameters and takes 1.5 hours to train, the cVAE module has 1.2 million parameters and takes 2.5 hours to train, and the PAdaIN block has 0.2 million parameters and takes 1 hour to train. The entire UDnet model has 2.6 million parameters and takes 5 hours to train.

In terms of computational complexity, the inference time for UDnet is 0.03 seconds per image on average, which is faster than some existing methods for underwater image enhancement.

In conclusion, our results suggest that UDnet is feasible from both an implementation and computational perspective. However, it is important to note that the computational requirements may vary depending on the size and complexity of the input images, as well as the hardware used for training and inference.

\subsection{Compared Methods}
To have a comprehensive and fair evaluation of our model, we compare it to 10 previous studies  including six conventional unsupervised methods (CLAHE \citep{CLAHE}, IBLA \citep{IBLA2017}, RGHS \citep{Huang2018Shallow-waterAcquisition}, UCM \citep{Iqbal2010EnhancingMethod}, UDCP \citep{Drews2013TransmissionImages}, ULAP \citep{Song2018ARestoration}) and four deep-learning-based methods (PIFM \citep{PIFM2021}, PUIEnet \citep{fu2022uiu}, USLN \citep{Xiao2022usln}, Wavenet \citep{Wavelength}).
The comparison with conventional  unsupervised methods aims to demonstrate the advantages of our trainable unsupervised deep-learning-based method.

We applied these conventional unsupervised approaches directly to the test sets. We used the respective studies' code and training approach for the deep learning-based methods. To guarantee the experiment's objectivity, we trained the four deep-learning-based methods on UIEBD \citep{Li2020AnBeyond} and applied the author-provided model and network training parameters.
 % % %%%%%%%%%%%%%%%%%%%%%%%%%%%%%%%%%%%%%%%%%%%%%%%%%%%%%%%%%%%%%%%%
\begin{figure*}[!ht]%[ht]
\centering
\includegraphics[width=0.85\textwidth]{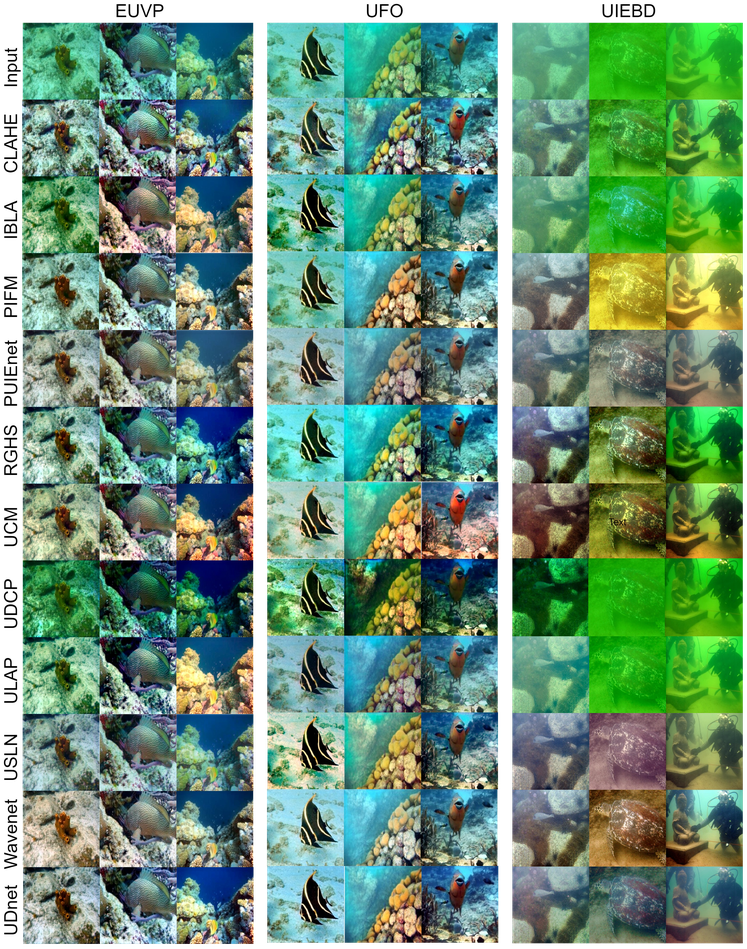}
\caption{Visual comparisons on challenging underwater images sampled from paired datasets, i.e. EUVP \citep{Islam2020FastPerception}, UFO \citep{Islam2022seu}, and UIEBD \citep{Li2020AnBeyond}. The name on the right of each row refers to the enhancement method used.}
\label{fig:3}
\end{figure*}
% % %%%%%%%%%%%%%%%%%%%%%%%%%%%%%%%%%%%%%%%%%%%%%%%%%%%%%%%%%%%%%%%%

% % %%%%%%%%%%%%%%%%%%%%%%%%%%%%%%%%%%%%%%%%%%%%%%%%%%%%%%%%%%%%%%%%
\begin{figure*}[!ht]%[ht]
\centering
\includegraphics[width=0.85\textwidth]{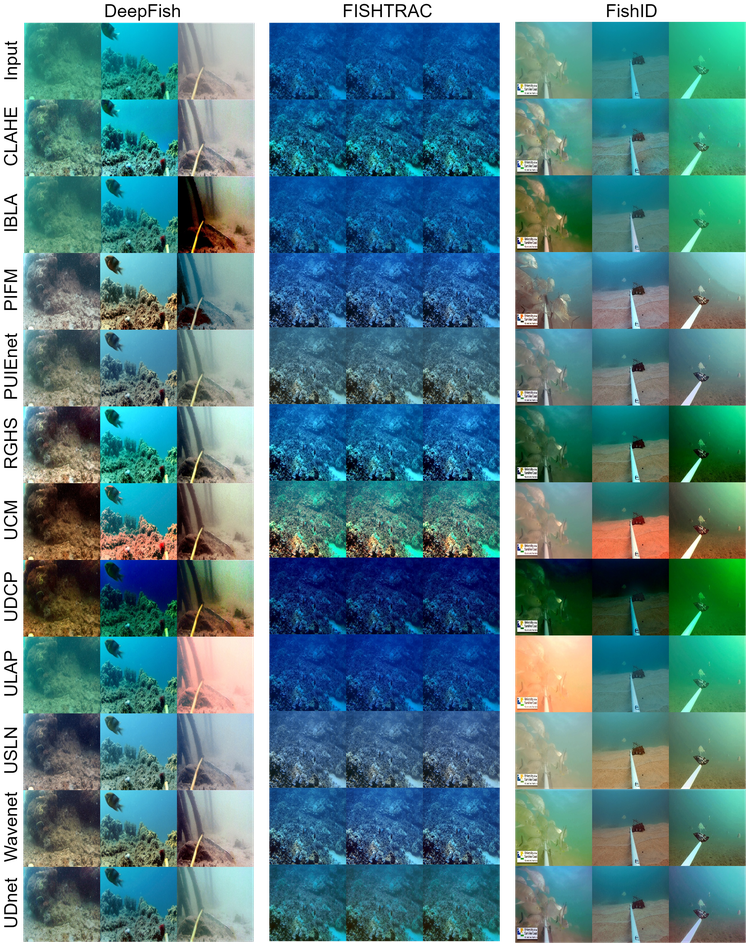}
\caption{Visual comparisons on challenging underwater images sampled from DeepFish \citep{Saleh2020}, FISHTRAC \citep{Mandel2023DetectionDetections}, and FishID \citep{lopiz2021eco}. The name on the right of each row refers to the method. We also include a short video of our model's prediction at \href{https://youtu.be/k4ASsGze5p8}{https://youtu.be/k4ASsGze5p8}
and \href{https://youtu.be/NV5GH-GG_3c}{https://youtu.be/NV5GH-GG\_3c}.}
\label{fig:4}
\end{figure*}
% % %%%%%%%%%%%%%%%%%%%%%%%%%%%%%%%%%%%%%%%%%%%%%%%%%%%%%%%%%%%%%%%%

% % %%%%%%%%%%%%%%%%%%%%%%%%%%%%%%%%%%%%%%%%%%%%%%%%%%%%%%%%%%%%%%%%
\begin{figure*}[!t]%[ht]
\centering
\includegraphics[width=0.91\textwidth]{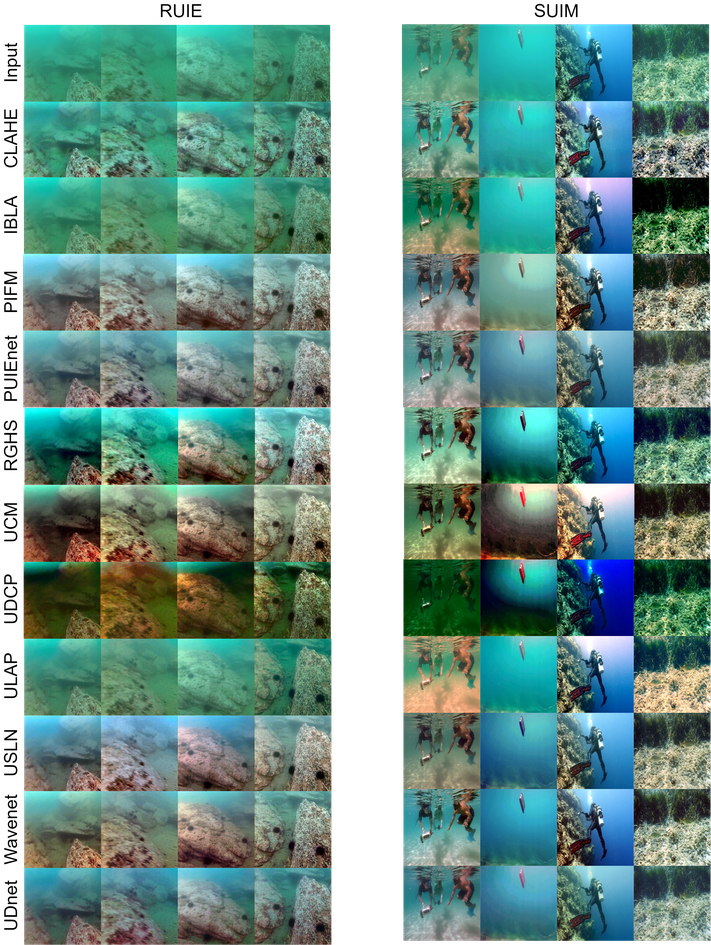}
\caption{Visual comparisons on challenging underwater images sampled from RUIE \citep{Liu2020Real-worldLight}, and SUIM \citep{Islam2020}. The name on the right of each row refers to the method. }
\label{fig:5}
\end{figure*}
% % %%%%%%%%%%%%%%%%%%%%%%%%%%%%%%%%%%%%%%%%%%%%%%%%%%%%%%%%%%%%%%%%

\subsection{Quantitative Comparisons} \label{quntres}
The comparison results for all paired test sets are summarized in  \cref{table1}. We report the average scores of the four full-reference metrics (PSNR, SSIM, MAD, GMSD).
 \cref{table1} demonstrates that our proposed method outperforms all six conventional unsupervised methods and all four deep-learning-based methods in all four full-reference metrics on the EUVP dataset and shows great performance on the UFO dataset.  Our model achieves the highest PSNR, SSIM scores on EUVP, and the lowest MAD, GMSD scores on EUVP, UFO.
In addition, we also found that 
\begin{itemize}
    \item  Although our model was trained in an unsupervised way, it still outperformed the fully supervised deep-learning-based models trained on UIEBD dataset on EUVP and UFO.
    \item This result shows that our proposed unsupervised deep-learning-based method is better than conventional ones at preserving structural information and contrast preservation, which suggests the superiority of our trainable model.
    
    \item Our model's performance is slightly lower than fully supervised methods on UIEBD dataset. However, this is because fully supervised methods were trained on the images and ground truth labels acquired from the UIEBD, while our method is fully unsupervised  and was not trained with ground truth labels. 
     
    \item Even without any extra labels, our model outperforms models that are trained on the UIEBD dataset on the two metrics of MAD and GMSD for paired datasets.
\end{itemize}

We also provided  quantitative comparisons for unpaired test sets in  \cref{table2}, which  demonstrate that our model achieves the highest NIQE on FISHTRAC, FishID, RUIE, SUIM, and the second-best UIQM score on DeepFish, FISHTRAC, SUIM. 
These results also show that 
\begin{itemize}
    \item Deep-learning-based models cannot outperform conventional approaches in no-reference evaluation metrics, in contrast to full-reference evaluation metrics.
    \item The quantitative results suggest that our method can generalize well on unseen datasets even without ground truth. 
\end{itemize}

Our model shows slight performance variation across different datasets and evaluation metrics in  \cref{table2}. This can be attributed to the inherent challenges of underwater image enhancement, including the variability of underwater conditions and the limited availability of high-quality training data. It is crucial to acknowledge that no-reference metrics, including UIQM, MUSIQ, and NIQE, may not consistently provide an accurate representation of image quality under certain circumstances. Consequently, the scores derived from these metrics are utilized solely as reference points within our study. Even despite the minor performance variation, our model still outperforms several studies and is on-par with the state-of-the-art.

 % was blue
According to the results presented in \cref{table5}, our proposed method, UDnet, demonstrates superior performance in underwater image enhancement when compared to other published works on four datasets: UIEBD, EUVP, UCCS and UIQS. Specifically, UDnet achieved the highest average values for PSNR, SSIM, UIQM and UCIQE metrics on these datasets. For instance, on the UIEBD dataset, UDnet achieved the highest UIQM and UCIQE values while on the EUVP dataset, it achieved the highest PSNR, UIQM and UCIQE values. Similarly, on the UCCS dataset, UDnet achieved the highest UIQM and UCIQE values and on the UIQS dataset, it achieved the highest UCIQE value. These results indicate that UDnet is a robust method for enhancing underwater images.

Our method, UDnet, performs better than other methods due to the effective combination of several components. The cVAE module is able to learn a compact and informative representation of the underwater image content, which helps to preserve important details during the enhancement process. The PAdaIN module is able to adaptively adjust the style of the enhanced image to match the target domain, resulting in more natural and visually pleasing results. Finally, the multi-colour space stretch module is able to effectively enhance the contrast and colour of the underwater images by stretching the colour histogram in multiple colour spaces. These components work together to produce high-quality enhanced underwater images.

% \subsection{Visual Comparisons}
\subsection{Qualitative Comparisons} \label{qultres}
Underwater images possess several unique characteristics. %They have more texture content and low luminance and contrast compared to natural images with more texture content and low luminance and contrast. 
They have more texture content and low luminance and contrast compared to terrestrial images.
Therefore, it is important to assess human visual perception in terms of image content enhancement in underwater images, especially in terms of colour enhancement. 
To gain more insight into the effectiveness of our proposed UDnet, we performed comprehensive investigations and comparisons among all eight data sets using the ten previous methods introduced. 

\cref{fig:3} demonstrates three example raw input images of each of the three paired datasets in the first row, along with the enhanced image outputs from the 10 aforementioned studies and our UDNet. 
%This figure shows a sample comparison between the output images generated by our model and those by the ten methods mentioned above.
This comparison has a two-fold purpose: 1) To demonstrate the effectiveness of the deep-learning-based methods in the no-reference settings. 2) To showcase the superiority of our unsupervised method, which has enhanced the underwater scenes without ground truth for training.

Furthermore, to prove the superiority of our model in handling unpaired images, we show visual comparisons of randomly selected underwater images from the five aforementioned unpaired datasets in \cref{fig:4}, and \cref{fig:5}. 
We also include a short video of our model's prediction at \href{https://youtu.be/k4ASsGze5p8}{https://youtu.be/k4ASsGze5p8}
and \href{https://youtu.be/NV5GH-GG_3c}{https://youtu.be/NV5GH-GG\_3c}. %Note that, these datasets are more challenging compared to the paired datasets.  

As \cref{fig:4} shows, the obvious light limitation of the raw image results in low contrast. For example, UDCP and ULAP models tend to make the image darker, while others such as UCMeven introduce reddish colour. In comparison, our model increases both brightness and contrast, making the details of the image clear. The input image samples given in \cref{fig:5} mostly suffer from obvious green deviation, which cannot be resolved by most models. For example, CLAHE, IBLA, and ULAP fail to remove the green deviation. In comparison, our model removes the greenish colour and makes the image colours balanced.

Overall, our qualitative comparison results show: 
\begin{itemize}
    \item Even when the ground truth label of the paired images is added to enhance visual quality, some of the previous methods show problems such as over-enhancement, lack of contrast, and saturation. 
    \item Some of the models' output images from the paired dataset have over- or under-enhanced backgrounds, while some have no change in the background. However, the output image of our model does not show such problems.
    \item Some of the models' output images' background pixels are saturated. However, our model has not suffered from the over- or under-saturation problem.
\end{itemize}

% % %%%%%%%%%%%%%%%%%%%%%%%%%%%%%%%%%%%%%%%%%%%%%%%%%%%%%%%%%%%%%%%%
\begin{figure*}[!t]%[ht]
\centering
\includegraphics[width=0.98\textwidth]{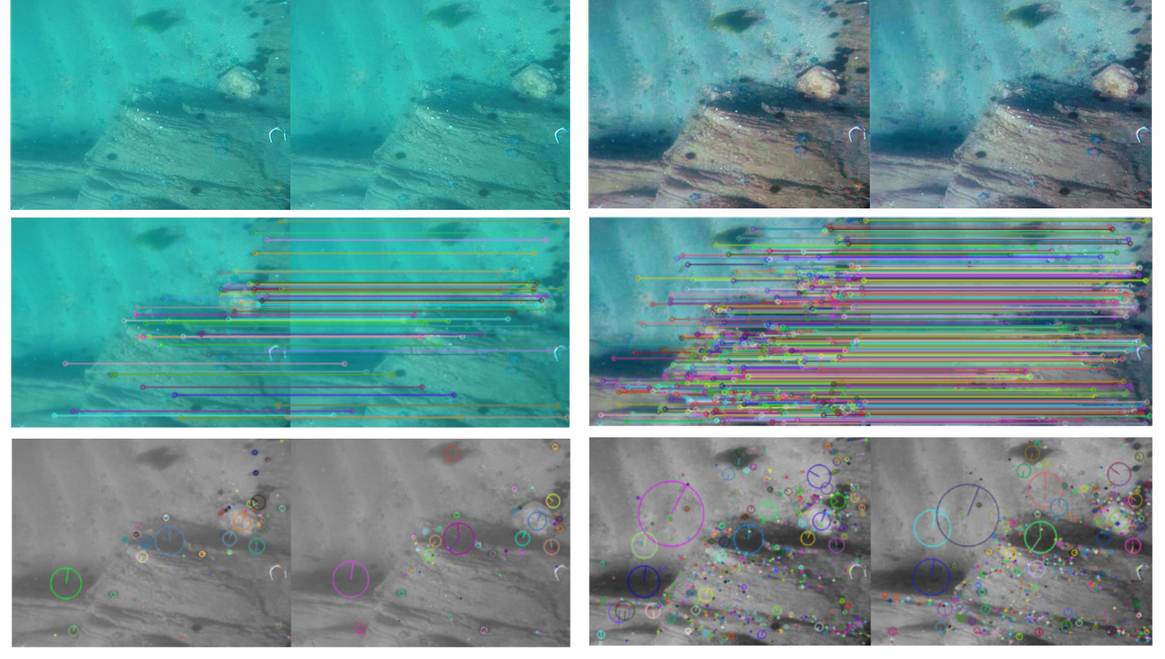}
\caption{Comparison of image feature and key points matching before (left) and after (right) image enhancement with our model. From the top: the original images, matched feature points, and SIFT keypoints. The images are from RUIE \citep{Liu2020Real-worldLight} dataset.}
\label{fig:2}
\end{figure*}
% % %%%%%%%%%%%%%%%%%%%%%%%%%%%%%%%%%%%%%%%%%%%%%%%%%%%%%%%%%%%%%%%%

\subsection{Visual Perception Improvement} \label{secvisper}
One of the main objectives of underwater image enhancement is to increase underwater robots' capacity to visually perceive their surroundings. This is essential for robots to make autonomous decisions in complex underwater scenarios. To evaluate our model's performance in visual perception improvement, we used feature detection and matching to assess its capability in improving the visual perception of underwater images. Feature detection and matching are commonly used techniques in many computer vision applications, such as structure-from-motion, image retrieval, object detection, and image stitching. Here, we use Scale-Invariant Feature Transform (SIFT), which helps locate the local features in an image (keypoints),
% \citep{low2004sift}, 
and Random Sample Consensus (RANSAC) \citep{liRANSAC}, which is used to match feature points. These methods are used to compare the visual perception of an underwater image before and after enhancement. 
\cref{fig:2} depicts the result for two consecutive frames from RUIE \citep{Liu2020Real-worldLight} dataset, (blue\_01.jpg) and (blue\_02.jpg). These show that the numbers of matched points between the two image frames increase from 74 (before the enhancement) to 594 after the enhancement. At the same time, the number of SIFT keypoints also dramatically increases as a result of the enhancement, significantly improving the visual perception of the environment.

% % %%%%%%%%%%%%%%%%%%%%%%%%%%%%%%%%%%%%%%%%%%%%%%%%%%%%%%%%%%%%%%%%

 % was blue
\subsection{Ablation Study} \label{secablation} To better understand how the proposed method works and what are the key factors that contribute to its performance, we conducted an ablation study to examine the impact of its different components and stages. These include the SGMCSS that adjusts the colour balance of the input images, the extra reference maps that are generated by applying different enhancement techniques to the input images, and the VGG loss that measures the perceptual similarity between the output images and the reference maps. We compared the full model with several ablated variants that remove or modify one of these components or stages. The quantitative comparisons are presented in \cref{table4}, where \begin{itemize} \item \textbf{w/o colour} means that UDnet is trained without using the SGMCSS in the reference map generation stage. The input images are directly fed to the cVAE without any colour adjustment. \item \textbf{All colour} means that UDnet is trained with applying the SGMCSS to all inputs, including the input images and the reference maps. This means that the colour balance of both the input images and the reference maps are adjusted before feeding them to the cVAE. \item \textbf{Multi-label} means that UDnet is trained with using 6 extra enhanced reference maps that are generated by applying different enhancement techniques to the input images, such as histogram equalization, CLAHE, and Retinex. This means that each input image has 9 reference maps in total, including the original 3 generated by contrast and saturation adjustment, and gamma correction. \item \textbf{w/o VGG} means that UDnet is trained without using VGG loss in the objective function. The model only minimizes the reconstruction loss between the output images and the reference maps. \end{itemize}

\begin{figure*}[ht]%[ht]
\centering
\includegraphics[width=0.99\textwidth]{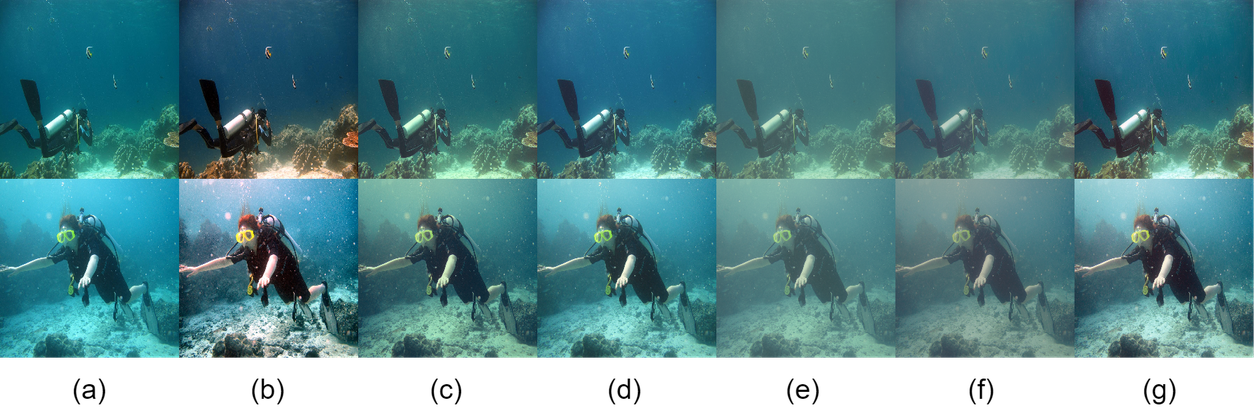}
\caption{ABLATION STUDY: The qualitative comparison of the contributions of multiple stages of the proposed framework on the UIEBD dataset. (a) Input, (b) ground truth, (c) w/o colour, (d) All colour, (e) Multi-label, (f) w/o VGG, (g) Full Model.}
\label{fig:6}
\end{figure*}

We used PSNR and SSIM to evaluate the results on UIEBD, which are shown in \cref{table4}. The results show that the entire model outperforms all variants ablated in both metrics, indicating that each component and stage of the proposed method is essential for achieving high-quality underwater image enhancement. The qualitative comparisons of the output images produced by different variants are presented in \cref{fig:6}. From these comparisons, we can draw the following conclusions:

%
%%%%%%%%%%%%%%%%%%%%%%%%%%%%%%%%%%%%%%%%%%%%%%%%%%%%%%%%%%%%%%%%
\begin{table}[!t]%[!t]%[h!]
\centering
\caption{ABLATION STUDY: COMPARISON AGAINST DIFFERENT MODEL VARIANTS ON UIEBD DATASET IN TERMS OF AVERAGE  PSNR AND SSIM VALUES}
\label{table4}
\small
\begin{spacing}{1.29}
\setlength{\tabcolsep}{22pt}
\begin{tabular}{lcc}
\hline
Model variant   & PSNR $\uparrow$ & SSIM $\uparrow$ \\
\hline
w/o colour      & 22.01    & 0.791    \\
All colour      & 21.73    & 0.784    \\
Multi-label     & 22.12    & 0.795   \\
w/o VGG         & 21.89    & 0.789    \\
\hline
Full Model      & 22.23    & 0.812    \\
\hline
\end{tabular}
\end{spacing}
\vspace{-1em}
\end{table}

%%%%%%%%%%%%%%%%%%%%%%%%%%%%%%%%%%%%%%%%%%%%%%%%%%%%%%%%%%%%%%%

% % %%%%%%%%%%%%%%%%%%%%%%%%%%%%%%%%%%%%%%%%%%%%%%%%%%%%%%%%%%%%%%%%
% % %%%%%%%%%%%%%%%%%%%%%%%%%%%%%%%%%%%%%%%%%%%%%%%%%%%%%%%%%%%%%%%%

\begin{enumerate} \item Without using the SGMCSS in the reference map generation stage, UDnet fails to produce satisfactory results (see \cref{fig:6}©). The output images still suffer from low contrast and poor visibility. For example, in the first row of \cref{fig:6}, it is hard to see the details of the sea bed and the green plants at the bottom of the image. This suggests that adjusting the colour balance of the input images is a crucial step for generating realistic reference maps that can guide UDnet to enhance underwater images. \item With applying the SGMCSS to all inputs, UDnet performs slightly better than without using it at all, but still worse than using it only in the reference map generation stage (see \cref{fig:6}(d)). The output images are brighter than w/o colour, but they also lose some details and colours. For example, in the second row of \cref{fig:6}, some parts of the fish are over-exposed and washed out. This implies that adjusting the colour balance of both the input images and the reference maps may introduce some inconsistency and distortion that can affect UDnet’s ability to learn from them. \item Adding more reference maps by using different enhancement techniques does not improve UDnet’s performance, but rather degrades it (see \cref{fig:6}(e)). The output images are over-enhanced and have unnatural colours. For example, in the third row of \cref{fig:6}, some parts of the coral are too bright and have a pinkish hue. This indicates that adding more reference maps does not necessarily provide more useful information for UDnet, but may introduce more noise and ambiguity that can confuse UDnet and make it harder to learn from them. \item When UDnet is trained without using VGG loss, the quality of the output images is significantly reduced (see \cref{fig:6}(f)). The output images have low contrast, poor visibility, and distorted colours. For example, in the fourth row of \cref{fig:6}, the output image is very dark and has a bluish tint. This demonstrates that VGG loss is an important component of the objective function that can help UDnet to learn more perceptual features and semantic information from the reference maps and improve the visual quality of the output images. \end{enumerate}

\begin{table}[ht]
\centering
\caption{Quantitative comparisons for the PAdaIN module and baseline model against the full UDnet model on the UIEBD dataset. Metrics include PSNR (dB) and SSIM. Higher values indicate better performance.}
\label{table6}
\begin{tabular}{lcc}
\hline
\textbf{Model Variant}      & \textbf{PSNR (dB)} & \textbf{SSIM} \\ \hline
Baseline Model              & 19.42              & 0.712         \\ 
w/o PAdaIN                 & 21.35              & 0.743         \\ 
Full Model (UDnet)         & \textbf{22.23}     & \textbf{0.812} \\ \hline
\end{tabular}
\end{table}

\noindent
\textbf{Effectiveness of the PAdaIN Module and Baseline Model Comparison:}

To further validate the significance of the PAdaIN module, we conducted additional experiments comparing its contribution to the performance of the proposed framework. Table~\ref{table6} presents the quantitative results of these experiments on the UIEBD dataset, using PSNR and SSIM as evaluation metrics. The results confirm that each component of the proposed method plays a crucial role in achieving superior performance.

Baseline Model: This variant, which excludes all proposed enhancements, achieves the lowest PSNR and SSIM values, highlighting the need for the advanced components integrated into the full UDnet model.
w/o PAdaIN: Removing the PAdaIN module causes a noticeable drop in performance compared to the full model, demonstrating that PAdaIN effectively encodes feature uncertainties and contributes to high-quality underwater image enhancement.
Full Model (UDnet): The complete model, including SGMCSS, cVAE, and PAdaIN modules, achieves the best results with a PSNR of 22.23 dB and SSIM of 0.812, highlighting the synergy of these components in the overall framework.
These findings underscore the importance of the PAdaIN module, particularly in refining the network’s ability to enhance underwater images.

\section{Discussion} \label{secdisc}
Enhancing underwater images is challenging due to the complex and diverse nature of underwater environments. Our proposed method, UDNet, addresses these challenges by adopting an unsupervised framework that leverages probabilistic uncertainty modeling during training. This novel approach enables UDNet to adaptively enhance underwater images with varying characteristics, setting it apart from traditional supervised methods that rely on large datasets of paired raw and enhanced images.

UDNet's key strengths include its unsupervised learning capability, which eliminates the need for ground truth data, and its innovative use of statistical information through the Statistically Guided Multi-Colour Space Stretch (SGMCSS) and Probabilistic Adaptive Instance Normalization (PAdaIN) modules. These modules improve robustness and enhance image quality by addressing variations in contrast, color balance, and illumination. Experimental results confirm UDNet's competitive performance across eight public datasets, demonstrating its ability to outperform or match state-of-the-art methods quantitatively and qualitatively.

Despite these advancements, UDNet has limitations that warrant further exploration. Backscatter, particularly at greater distances, remains a significant challenge, as it affects the visual clarity of enhanced images. While our approach mitigates some of these issues, its reliance on statistical models can occasionally result in unrealistic enhancements. Additionally, the environmental variability of underwater settings—ranging from oceans to lakes—means that the model's generalization may not always produce optimal results across all scenarios.

To address these limitations, future work will focus on enhancing UDNet's robustness by exploring alternative CNN architectures and integrating multi-resolution approaches to capture finer details. Improved methods for reference map generation could also further reduce dependence on statistical assumptions, resulting in higher-quality enhancements.

The potential applications of UDNet are vast. Enhanced underwater images can significantly benefit environmental monitoring, providing insights into marine ecosystems and supporting conservation efforts. In marine biology, UDNet's improvements can aid in the study of species behavior and habitats. Moreover, in underwater archaeology, the model's ability to clarify images can facilitate the study of submerged artifacts and structures.

Beyond its current scope, UDNet has the potential to serve as a framework for generating high-quality reference maps for other domains, such as medical imaging or satellite image enhancement, where ground truth data is challenging to obtain.

\section{Conclusion} \label{secconc}

In this work, we introduced UDNet, an unsupervised deep learning framework designed for underwater image enhancement. By leveraging probabilistic uncertainty modeling and an encoder-decoder architecture, UDNet effectively addresses challenges such as random distortion and low contrast inherent in underwater images. Its innovative design—featuring the SGMCSS and PAdaIN modules—enables robust image enhancement without relying on manually labeled data, marking a significant advancement in the field.
Our experimental results demonstrate that UDNet outperforms ten state-of-the-art underwater image enhancement methods across seven metrics and eight datasets, underscoring its versatility and effectiveness. UDNet's strong generalization ability, particularly with unpaired datasets, positions it as a practical tool for diverse underwater applications.

%%%%%%%%%%%%%%%%%%%%%%%%
\section*{Acknowledgement}
This research is supported by the Australian Research Training Program (RTP) Scholarship and Food Agility HDR Top-Up Scholarship. D. Jerry and M. Rahimi Azghadi acknowledge the Australian Research Council through their Industrial Transformation Research Hub program.  

% \section*{Funding}
% This project was supported by CRC project funding from the Department of Industry, Innovation and Science, Commonwealth of Australia. Mainstream Aquaculture Group partnered with James Cook University, The University of Melbourne and four commercial farming operations to deliver this project. 

%%%%%%%%%%%%%%%%%%%%%%%%
% \clearpage	
% **** bibliography**** 
% \/ \/ \/ \/ 
\ifCLASSOPTIONcaptionsoff
\newpage
\fi
% \IEEEtriggeratref{8}

% \bibliographystyle{IEEEtran}
% % \bibliographystyle{IEEEtranN}
% \bibliography{thereferences}
	
% \bibliographystyle{apacite} % Specify APA citation style
\bibliography{thereferences} % Reference to your .bib file
	
\end{document}